\documentclass[letterpaper, 10 pt, conference]{ieeeconf}

\usepackage{graphicx}
\usepackage{xcolor}
\usepackage[hidelinks]{hyperref}
\usepackage{amsmath}
\usepackage{amssymb}
\usepackage{mathtools}
\usepackage{bm}
\usepackage{glossaries}
\usepackage{caption}
\usepackage{siunitx}
\usepackage{subcaption}
\usepackage{booktabs}
\IEEEoverridecommandlockouts
\newacronym{humvib}{HUMVIB}{\textit{HUMan-structure interaction and gait adaptation during locomotion on VIBrating structures}}
\newacronym{hsi}{HSI}{Human-Structure Interaction}
\newacronym{ppo}{PPO}{Proximal Policy Optimization}
\newacronym{rl}{RL}{Reinforcement Learning}
\newacronym{drl}{DRL}{Deep Reinforcement Learning}
\newacronym{com}{CoM}{Center of Mass}

\newcommand{\bbone}{\text{\usefont{U}{bbold}{m}{n}1}}
\MakeRobust{\bbone}

\title{\LARGE \bf
Bridge the Gap:\\
Enhancing Quadruped Locomotion with Vertical Ground Perturbations
}

\author{
Maximilian Stasica$^{*,1}$, Arne Bick$^{*,1,2}$, Nico Bohlinger$^{*,2}$, Omid Mohseni$^{1,3}$, Max Johannes Alois Fritzsche$^{4}$, \\ Clemens Hübler$^{4}$, Jan Peters$^{2,5}$, and André Seyfarth$^{1}$
\thanks{This work was funded by German Research Foundation (DFG) under the HUMVIB project (grant number: 446124066) and the INTENTION project (grant number: 506123304).
This project was also partially funded by National Science Centre, Poland under the OPUS call in the Weave program UMO-2021/43/I/ST6/02711.}%
\thanks{$^{*}$Equal contribution.}%
\thanks{$^{1}$M. Stasica, A. Bick, O. Mohseni and A. Seyfarth are with the Lauflabor Locomotion Laboratory, Institute of Sports Science and Centre of Cognitive Science,
        Technical University of Darmstadt, 64289 Darmstadt, Germany
        {\tt\small stasica@sport.tu-darmstadt.de}}%
\thanks{$^{2}$A. Bick, N. Bohlinger and J. Peters are with  the Intelligent Autonomous Systems group, Department of Computer Science, Technical University of Darmstadt,
        64289 Darmstadt, Germany}%
\thanks{$^{3}$O. Mohseni is with the Measurement and Sensor Technology Group, Department of Electrical Engineering and Information Technology, Technical University of Darmstadt, Darmstadt, Germany}%
\thanks{$^{4}$M. J. A. Fritzsche and C. Hübler are with the Institute of Structural Mechanics and Design, Technical University of Darmstadt,
        64289 Darmstadt, Germany}%
\thanks{$^{5}$J. Peters is with the German Research Center for AI (DFKI), Research Department: Systems AI for Robot Learning; Hessian.AI; Robotics Institute Germany (RIG)}%
}

\usepackage{color,soul} 
\begin{document}

\maketitle
\thispagestyle{empty}
\pagestyle{empty}

\begin{abstract}

Legged robots, particularly quadrupeds, excel at navigating rough terrains, yet their performance under vertical ground perturbations, such as those from oscillating surfaces, remains underexplored.
This study introduces a novel approach to enhance quadruped locomotion robustness by training the Unitree Go2 robot on an oscillating bridge---a 13.24-meter steel-and-concrete structure with a 2.0 Hz eigenfrequency designed to perturb locomotion.
Using Reinforcement Learning (RL) with the Proximal Policy Optimization (PPO) algorithm in a MuJoCo simulation, we trained 15 distinct locomotion policies, combining five gaits (trot, pace, bound, free, default) with three training conditions: rigid bridge and two oscillating bridge setups with differing height regulation strategies (relative to bridge surface or ground).
Domain randomization ensured zero-shot transfer to the real-world bridge.
Our results demonstrate that policies trained on the oscillating bridge exhibit superior stability and adaptability compared to those trained on rigid surfaces.
Our framework enables robust gait patterns even without prior bridge exposure.
These findings highlight the potential of simulation-based RL to improve quadruped locomotion during dynamic ground perturbations, offering insights for designing robots capable of traversing vibrating environments.

\end{abstract}

\begin{figure}[t]
    \centering
    \includegraphics[width=\linewidth]{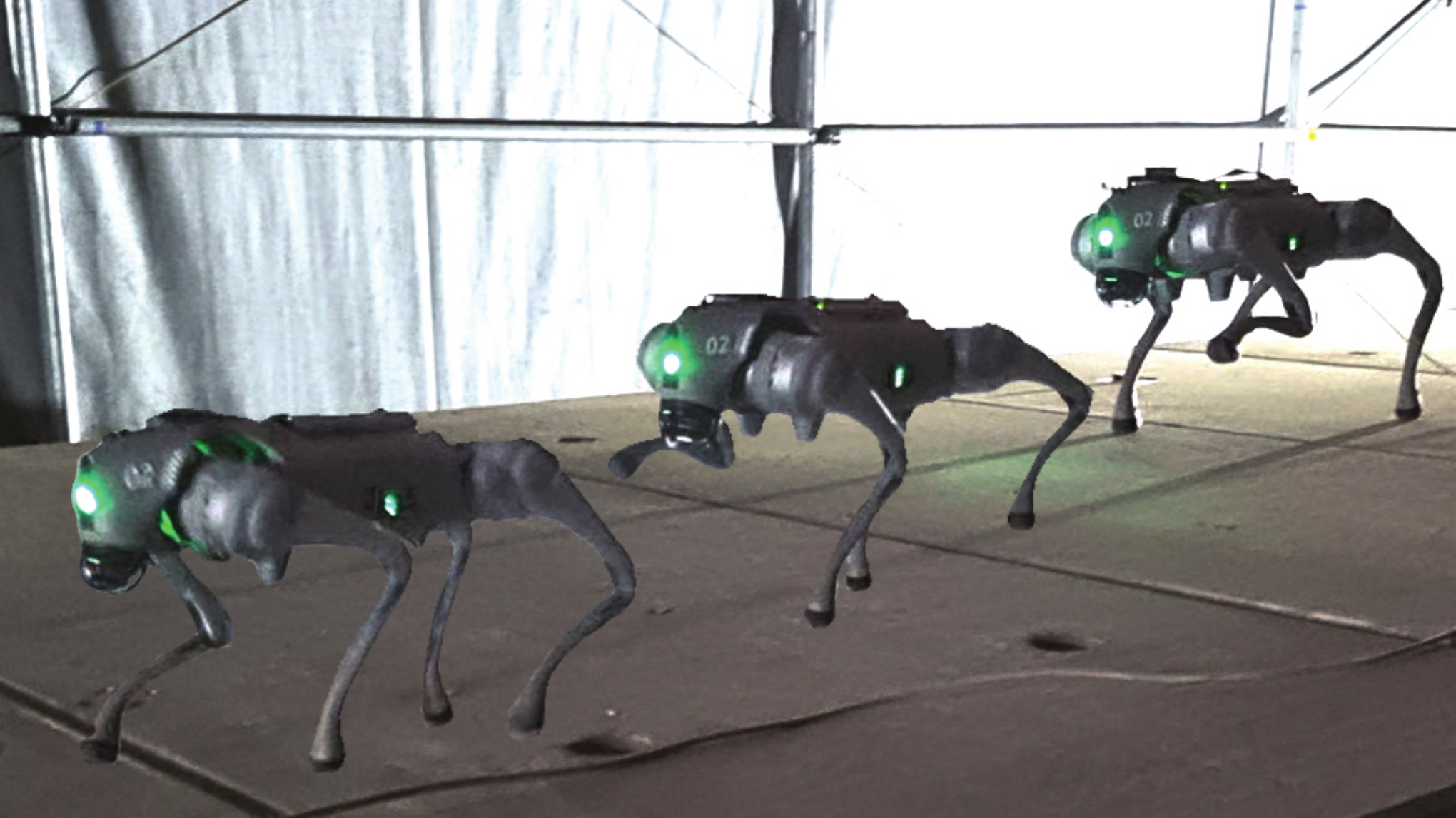}
    \caption{
        We train the Unitree Go2 quadruped in simulation and zero-shot transfer the learned policies to the oscillating real-world \glsfirst{humvib} bridge to investigate the impact of vertical ground perturbations on locomotion robustness.
    }
    \label{fig.real_robot_multi}
\end{figure}

\section{INTRODUCTION}
The development of legged robots, particularly quadrupeds, has surged in popularity due to their ability to traverse challenging terrains such as mountainsides \cite{miki2022learning} and obstacle-rich environments \cite{cheng2023, hoeller2024anymal}.
While these approaches can handle rigid uneven surfaces, their performance under active ground perturbations---encompassing moving obstacles, and both horizontal and vertical ground movements---remains under-examined \cite{ha2024learning}.
Although many robot designs and controllers have been evaluated on terrain with different stiffness levels \cite{choi2023learning} or friction coefficients \cite{ji2022concurrent, margolis2024rapid}, vertical ground movements are rarely studied, despite their relevance to real-world scenarios such as disaster zones, industrial sites, or hazardous fields.
This gap, which also appears in human locomotion research \cite{tokur2020review}, arises partly from the challenge of replicating such dynamic conditions in controlled environments---leaving a critical blind spot in our understanding of robotic adaptability to unstable surfaces.

Traditional robotic controllers often struggle with dynamic perturbations, as they are typically designed for predictable or rigid terrains.
While some systems manage uneven ground or obstacles effectively \cite{cheng2023, hoeller2024anymal}, their reliance on pre-tuned parameters limits resilience to sudden vertical shifts.
In human studies, vertical perturbations like varying ground stiffness significantly alter performance.
Human runners adapt instantly to these perturbations by adjusting leg mechanics \cite{mcmahon1978fast, ferris1999runners}, but the robotic equivalent remains under-explored.
The logistical challenge of testing such scenarios has hindered progress, underscoring the need for both innovative platforms and advanced control strategies to enhance locomotion robustness in unpredictable settings.

In recent years, \gls{drl} has become a popular approach for learning agile locomotion controllers for legged robots \cite{ha2024learning, bohlinger2025gait}.
On-policy \gls{drl} algorithms, such as \gls{ppo} \cite{schulman2017proximal}, can be scaled up through many parallel simulated physics environments and big batch sizes to learn complex and dynamic locomotion behaviors for any legged robot form factor \cite{rudin2022, ji2022concurrent, bohlinger2024onepolicy, ai2025towards}.
There are many examples of learning highly agile and robust locomotion for quadruped robots, such as fast-paced running and turning \cite{margolis2024rapid, mahankali2024maximizing}, dynamic jumping on parkour courses \cite{caluwaerts2023barkour, zhuang2023, cheng2023, chane2024soloparkour}, climbing on ladders or over tall obstacles \cite{vogel2024robust, hoeller2024anymal}, hiking on mountainous terrain \cite{miki2022learning} and even performing handstands, backflips and other acrobatic maneuvers \cite{cheng2023, kim2024stage}.
To transfer the learned policies to real-world robots, a wide range of domain randomization is necessary to bridge the sim-to-real gap \cite{tobin2017domain, rudin2022, ji2022concurrent, caluwaerts2023barkour}.
This includes randomizing the robot's properties, such as mass, inertia, and actuator dynamics, as well as the environment's properties, such as ground friction and roughness of the terrain.
To further improve the robustness and to teach the policy to adapt to unforeseen external disturbances, the training has to be augmented with adversarial perturbations that can occur at any moment in a learning episode.
However, the majority of works only consider sensor noise and pushes on the robot's trunk as active disturbances \cite{rudin2022, vogel2024robust, shi2024rethinking}, while the effects of active ground perturbations, like vertical ground movements, on the robot's locomotion are widely unexplored.

\begin{figure}[t]
    \centering
    \includegraphics[width=\linewidth]{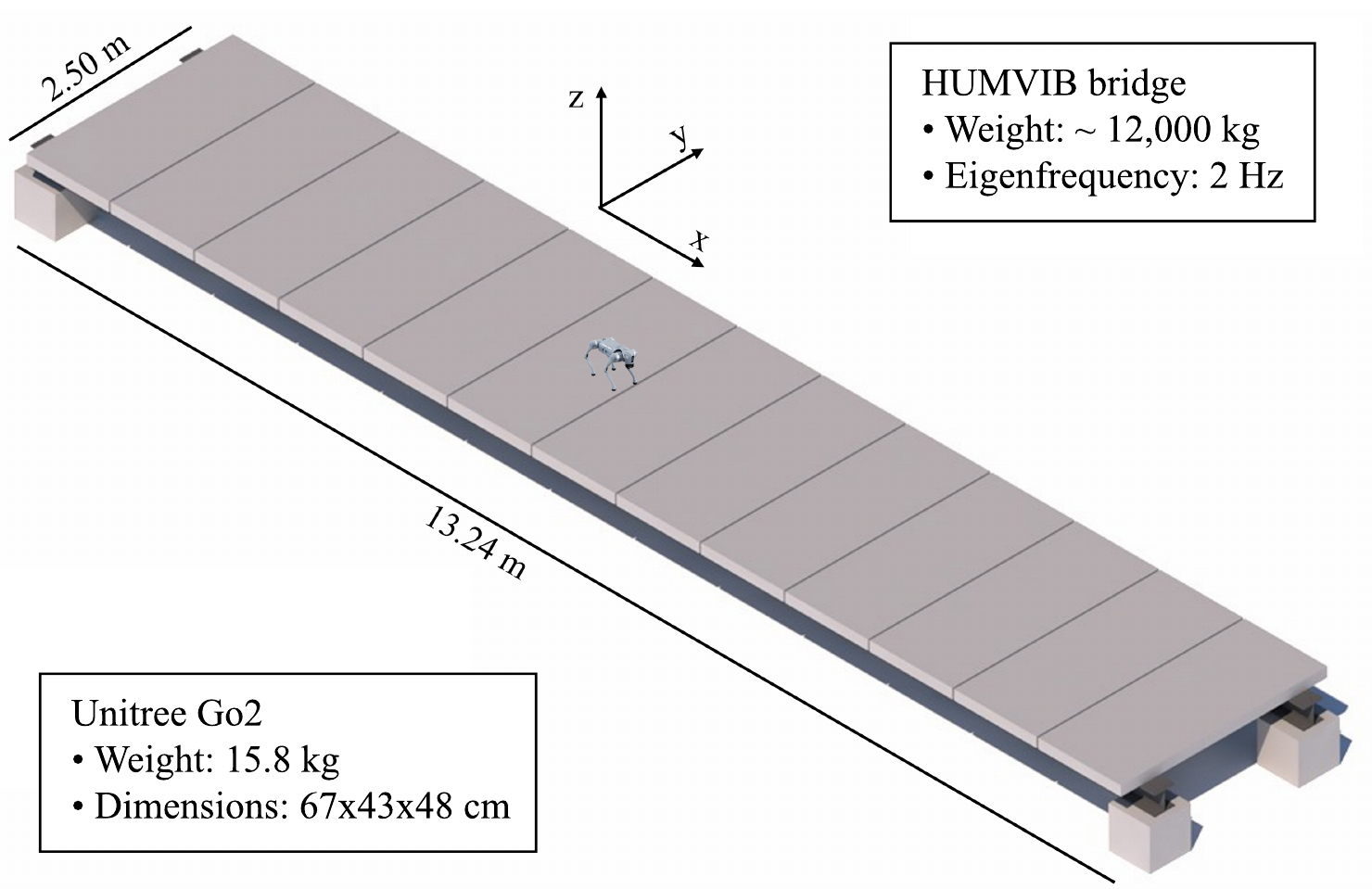}
    \caption{
        Schematic of the Unitree Go2 quadruped crossing the 13.24-meter long \glsfirst{humvib} bridge.
    }
    \label{fig.bridge_sketch}
\end{figure}

Understanding how \gls{rl} policies shape locomotion requires examining quadruped gaits, which are defined by the coordination of leg movements.
Common quadruped gaits include the trot (diagonal legs move together), pace (lateral legs move in unison), bound (front and back legs pair for a leaping motion), and more flexible styles like free (unconstrained coordination) \cite{fukuoka2015simple}.
The interplay between gait selection and dynamic ground conditions, such as vertical oscillations, still have to be fully understood in robotics.
Investigating these gait types under such perturbations is crucial for assessing how robots maintain stability and robustness, complementing the focus of \gls{rl} on policy optimization with insights into physical locomotion dynamics.

To tackle these challenges, we introduce a novel approach using the Unitree Go2 quadruped (Unitree, Hangzhou, China) on a purpose-built oscillating bridge (\autoref{fig.bridge_sketch}).
As part of the \gls{humvib} project, this structure, with a span of \SI{13.24}{\meter}, composed of two steel beams, and thirteen concrete slabs, features an eigenfrequency of approximately \SI{2}{\hertz}, making it highly susceptible to locomotion-induced oscillations \cite{firus2016tu, fritzsche2022integrated}.
Equipped with a \SI{2.5}{\meter} wide track and sensors like force plates and accelerometers, the bridge provides a controlled yet dynamic testbed for vertical perturbations.
By training the robot in simulation with state-of-the-art \gls{rl} techniques and evaluating its gaits in the real-world setting \cite{stasica2025bridge}\footnote{We provide code and videos of our real-world experiments at: \url{https://nico-bohlinger.github.io/bridge_the_gap_website/}.}, our study bridges the gap between rigid terrain research and the demands of unstable environments, aiming to enhance robotic locomotion stability and adaptability.

\section{LEARNING LOCOMOTION ON AN OSCILLATING BRIDGE}
We propose learning locomotion policies for quadruped robots on an oscillating bridge to investigate the impact of vertical ground perturbations on the robot's locomotion and achieve higher robustness of the learned policies.
We first describe the \gls{rl} training setup and the modeling of the bridge in simulation, followed by the necessary components for learning different gaits on the bridge, and the evaluation setup in the real world.

\subsection{Training setup}
We trained the Unitree Go2 quadruped in simulation using the CPU-based MuJoCo physics engine \cite{todorov2012mujoco} with 48 parallel environments for fast data collection.
Like previous works \cite{rudin2022, ji2022concurrent,bohlinger2024onepolicy}, we used the \gls{ppo} algorithm \cite{schulman2017proximal} to learn different locomotion policies.
We built on the \gls{drl} library RL-X \cite{bohlinger2023rlx} to implement and integrate the algorithm with the simulation environment.
The policies were trained to control the robot at \SI{50}{\hertz} with target joint positions, and to walk with a given x-y-yaw-command velocity $\bar{v} \in [-1.0, 1.0]^3$ with respect to the robot's trunk.
To enable zero-shot transfer of the learned policies to the real robot, we applied a wide range of domain randomization during training, including randomizing the robot's mass, inertia, \gls{com} (CoM), actuator dynamics and delays, the ground properties, such as friction and compliance, sensor noise, and pushes on the robot's trunk.
To investigate the robustness of the policies through the impact of vertical ground perturbations with the \gls{humvib} bridge, we trained the robot to walk on either a rigid or harmonically oscillating ground with varying eigenfrequencies and amplitudes.

\subsection{Bridge model}
The \gls{humvib} bridge is modeled in MuJoCo as a harmonic oscillator emulating the dynamics of the real bridge.
We fixed the surface of the bridge at \SI{1.05}{\meter} over the ground ---the peak of the oscillation---while the equilibrium position can be adjusted to modify the oscillation amplitude.
The stiffness is tuned such that the bridge exhibits an eigenfrequency of \SI{2}{\hertz} with an oscillation amplitude of $\pm$\SI{0.1}{\meter}, similar to the real bridge.
While this is a simplification of the bridge's real world behaviour, it accurately captures the structural response in the middle section.
The length of the simulated structure was selected to match the real-world setup.
During training, we varied the eigenfrequency of the bridge between \SI{0.75}{\hertz} and \SI{7.5}{\hertz} and the amplitude between zero and a constrained maximum value that ensures the bridge's acceleration remains below \SI{9.81}{\meter/\square\second} given its mass and adjusted stiffness.
This constraint is necessary to prevent the robot from experiencing too much acceleration to become airborne. 
Due to the high total mass of the bridge, the influence of the robot's mass is negligible and was omitted in the simulation.

\subsection{Learning different gaits}
We used the same reward terms, coefficients and curriculum as in \cite{bohlinger2024onepolicy}, and denoted the learned gait style as \textit{default}.
To encourage the emergence of the different \textit{trot}, \textit{pace}, and \textit{bound} gaits, we modified the existing symmetry reward term to penalize deviations from the characteristic stance phases of the respective gait.
Finally, removing the symmetry constraints results in the gait style we refer to as \textit{free}.
All reward terms can be found in appendix \ref{app:reward_function}.

Beyond the gait-specific reward terms, the environments in which the policies are learned can significantly shape the resulting gaits.
For gaits learned on the oscillating bridge, two different policies were trained by altering the base height reward term, which promotes the robot to maintain a constant height of \SI{0.325}{\meter}---the trunk height in the nominal standing position---with respect to the surface it is walking on.
In one case, the robot is rewarded to maintain a constant height relative to the oscillating bridge surface, called \textit{equidistant bridge} (\textit{eb}), while in the other, it is encouraged to maintain a constant height relative to the ground, called \textit{equidistant ground} (\textit{eg}).
When training on the rigid bridge, the base height reward term only has a single interpretation as the robot's height with respect to the ground, called \textit{no oscillation} (\textit{nos}).
Overall, this approach results in 15 distinct policies, derived from the five different gaits (\textit{trot}, \textit{pace}, \textit{bound}, \textit{free}, \textit{default}), each trained under three conditions, one on the rigid bridge and two on the oscillating bridge with the different height regulation strategies (\textit{nos}, \textit{eb}, \textit{eg}) respectively.

\subsection{Real-world setup}
The real world experiments took place on the \gls{humvib} bridge. 
To validate the presented algorithm, the structure was equipped with six Delsys Trigno Wireless sensors (Delsys, Natick, US) which were used as an IMU to track the bridges acceleration in different locations.
For each combination of gait style and training setting, the robot had to complete eight passes over the pre-oscillated and the idle bridge (\autoref{fig.real_robot_multi}) respectively.
The command velocity $\bar{v}$ was controlled by a human operator.
This ensured a constant speed of \SI{0.5}{\meter/\second} in the x-direction and the operator could adjust the y- and yaw-velocity to keep the robot on track in the case of lateral drift.

\section{RESULTS}

\begin{figure}[t]
    \centering
    \includegraphics[width=\linewidth]{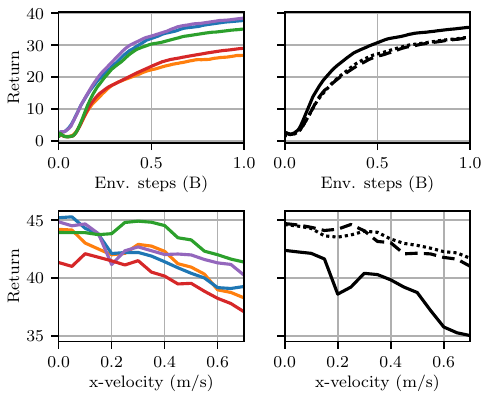}
    \includegraphics[width=\linewidth]{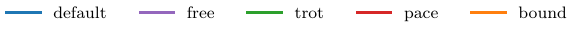}
    \includegraphics[width=0.9\linewidth]{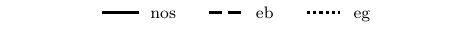}
    \caption{
    Top: Episode return for the different gaits (left) and training conditions (right) during training.
    Bottom: Episode return over the command velocity on the oscillating bridge for the different gaits (left) and training conditions (right) of the final policies during evaluation.
    }
    \label{fig.episode_returns}
\end{figure}

We first evaluated the learning dynamics of the different gaits and training conditions.
To ensure a fair comparison, we measured the performance of the policies by calculating the episode return without the gait-specific reward terms.
\autoref{fig.episode_returns} shows that the \textit{free} and \textit{default} gaits are the easiest to learn, as they impose no or fewer restrictions on the policy compared to the \textit{trot}, \textit{pace}, and \textit{bound} gaits that rely on strong gait-specific reward terms to enforce their desired footfall patterns.
We show all of the learned footfall patterns in \autoref{fig.footfall} and provide further analysis of the learned gaits in appendix \ref{app:gait_percentage}.
Of the three gait styles, the \textit{trot} gait is visually and reward-wise the closest to the \textit{default} gait, while the \textit{pace} and \textit{bound} gaits are more distinct and achieve significantly lower returns.
This indicates that these policies are worse at tracking the command velocity and more susceptible to disturbances from the training environment.
When comparing the training conditions, the \textit{nos} policy performs best, as it is trained on the rigid surface and is less disturbed by the oscillations, while the \textit{eb} and \textit{eg} policy perform similarly to each other.

\begin{figure}[t]
    \centering
    \includegraphics[width=\linewidth]{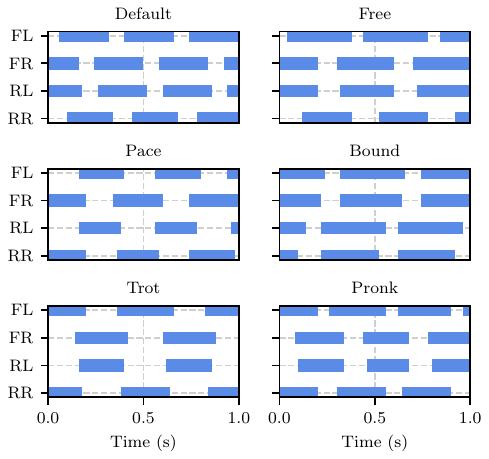}
    \caption{
        Footfall pattern of the different gait styles on the rigid bridge with a target speed $v_x = 0.5$\,m/s.
        The feet of the robot are denoted with front left (\textit{FL}), front right (\textit{FR}), rear left (\textit{RL}), and rear right (\textit{RR}).
        In our study, we omit the \textit{pronk} gait, as we were unable to train a policy that sufficiently learned the desired footfall pattern, due to the agent resorting to a more stable trot-like gait during learning.
    }
    \label{fig.footfall}
\end{figure}

Next, we evaluated the performance of the final policies on the oscillating bridge with varying command velocities.
\autoref{fig.episode_returns} shows that the \textit{free} and \textit{default} gaits perform best with smaller command velocities, where most of the probability mass for the sampled commands is located.
The \textit{trot} gait performs well at higher command velocities, and even outperforms the \textit{free} and \textit{default} gaits.
The \textit{pace} and \textit{bound} gaits perform worse than the other gaits in most cases.
When comparing the training conditions, the \textit{nos} policy performs significantly worse on the oscillating bridge than the \textit{eb} and \textit{eg} policies, highlighting the importance of training locomotion policies with vertical ground perturbations to achieve higher robustness in such environments.

\begin{figure}[t]
    \centering
    \includegraphics[width=\linewidth]{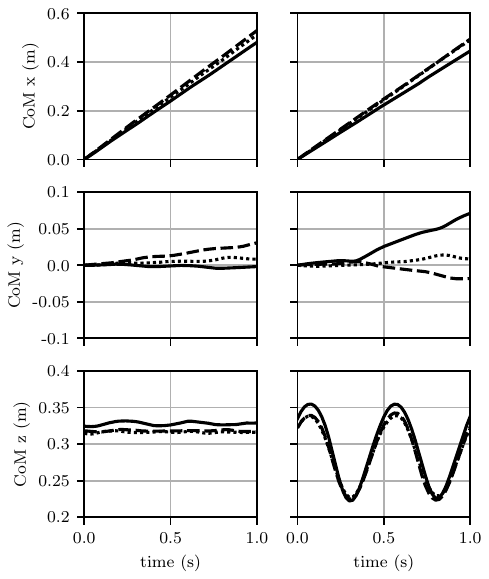}
    \includegraphics[width=\linewidth]{images/styles_legend.pdf}
    \caption{
        Movement of the robot's \gls{com} with a target speed $v_x = 0.5$\,m/s for the \textit{default} gait pattern under three conditions: no oscillation (\textit{nos}), equidistant bridge (\textit{eb}), and equidistant ground (\textit{eg}).
        Subfigures on the first column show x, y, and z directions of robot's \gls{com} on an idle bridge, while the subfigures on the second column show the same quantities, but on an oscillating bridge.
    }
    \label{fig.CoM_all}
\end{figure}

\begin{figure*}
    \centering
    \begin{subfigure}{0.49\textwidth}
        \centering
        \includegraphics[width = 0.3\textwidth]{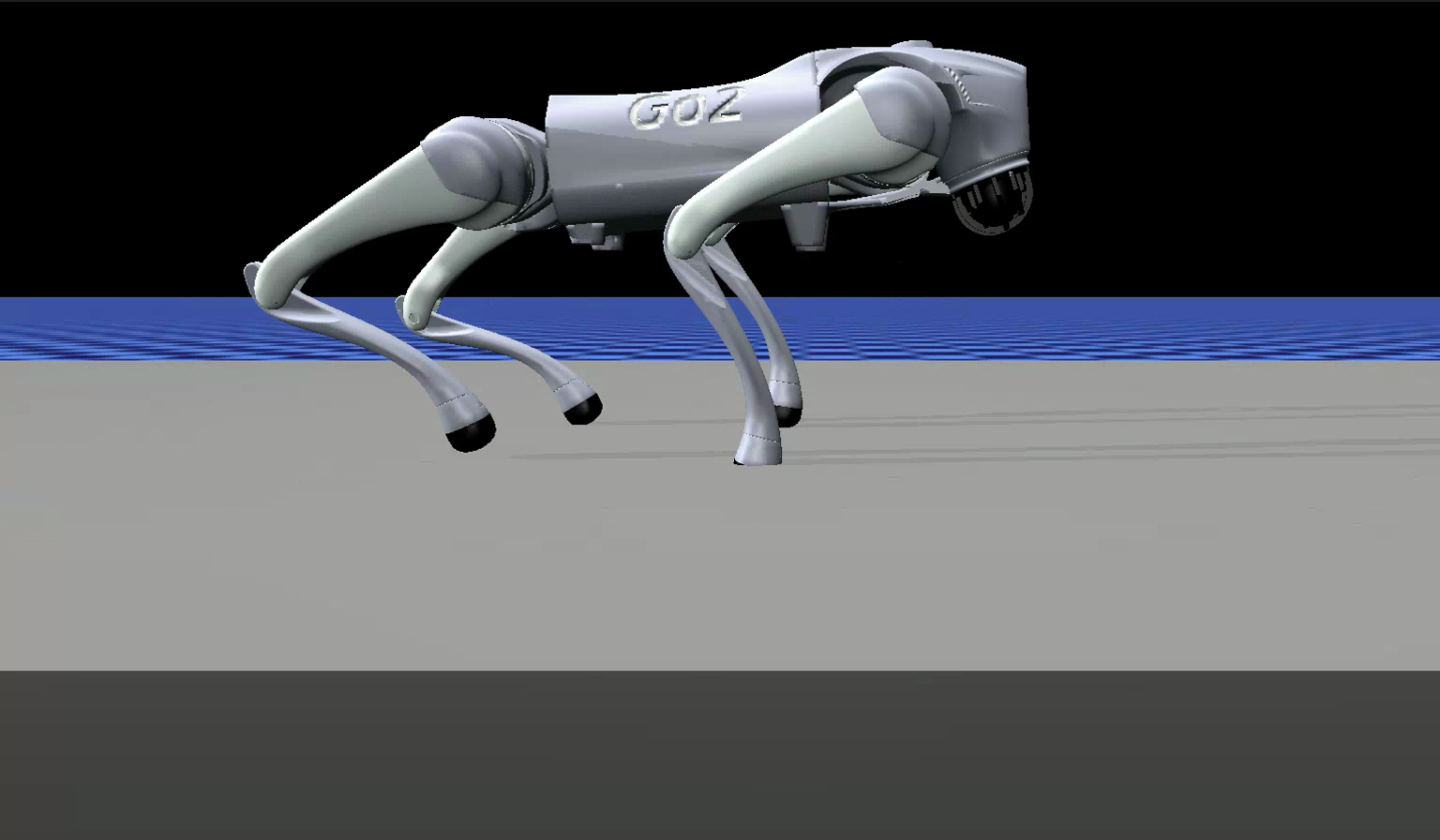}%
        \includegraphics[width = 0.3\textwidth]{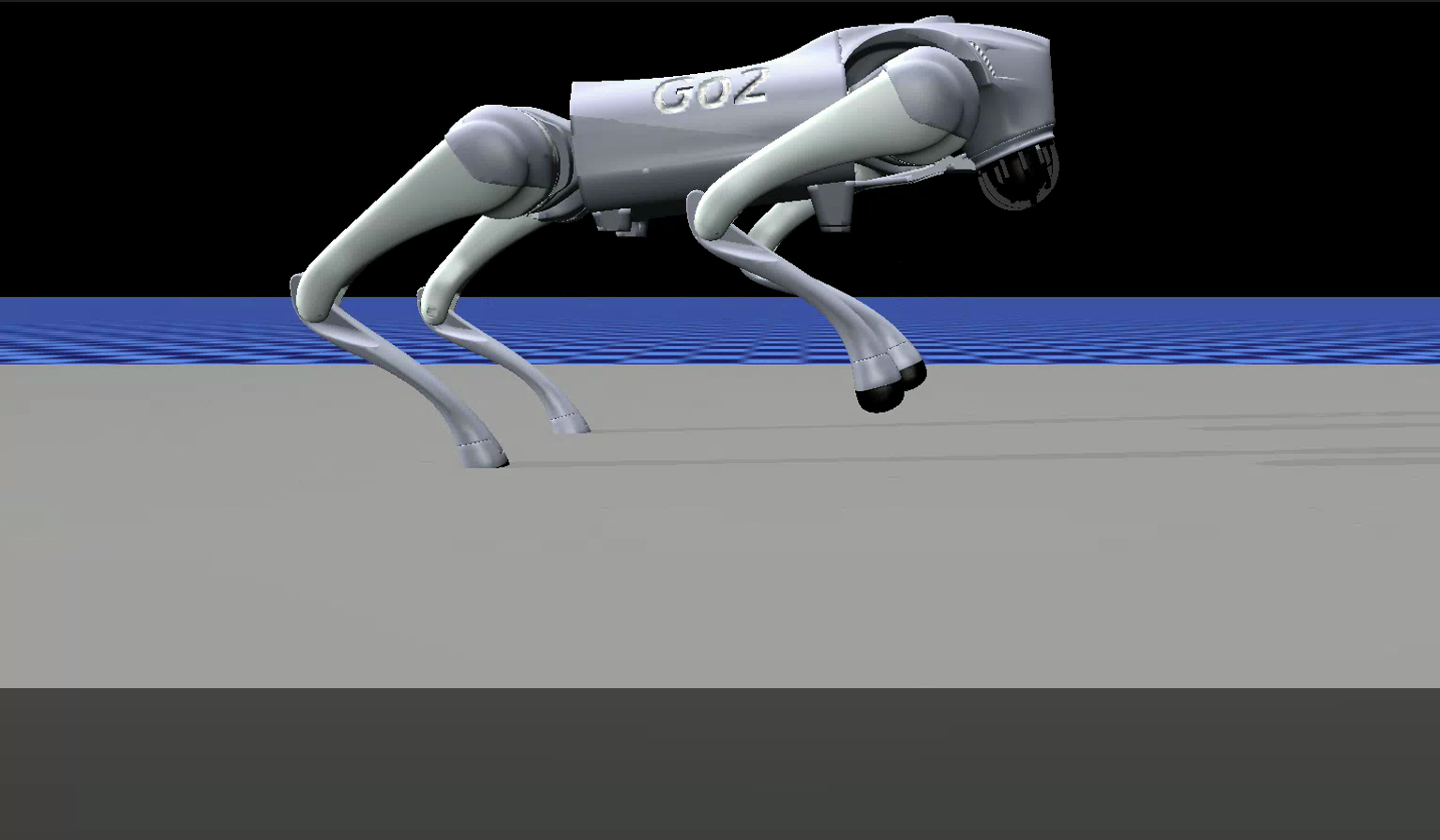}%
        \includegraphics[width = 0.3\textwidth]{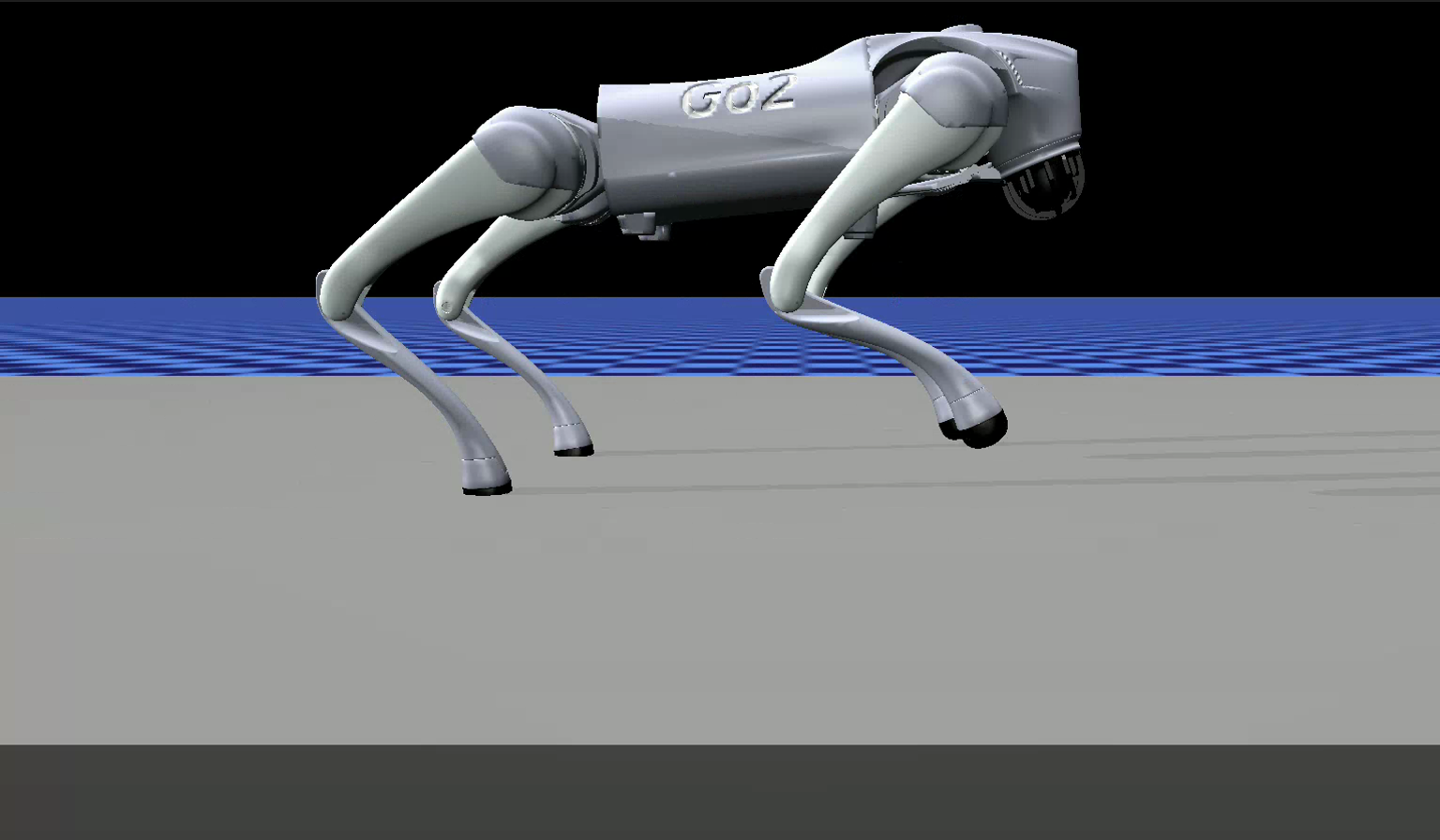}%
        \\
        \includegraphics[width = 0.3\textwidth]{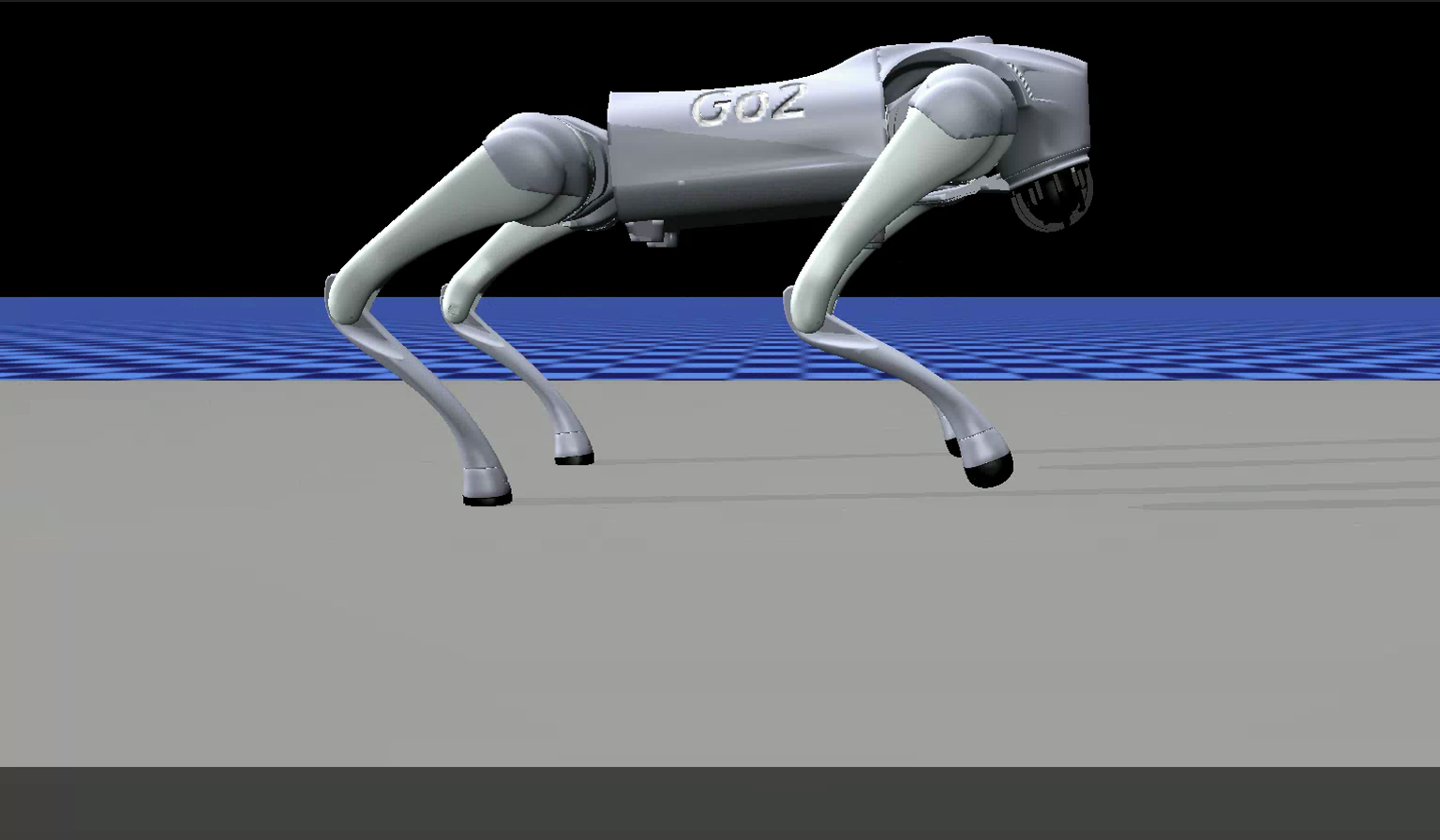}%
        \includegraphics[width = 0.3\textwidth]{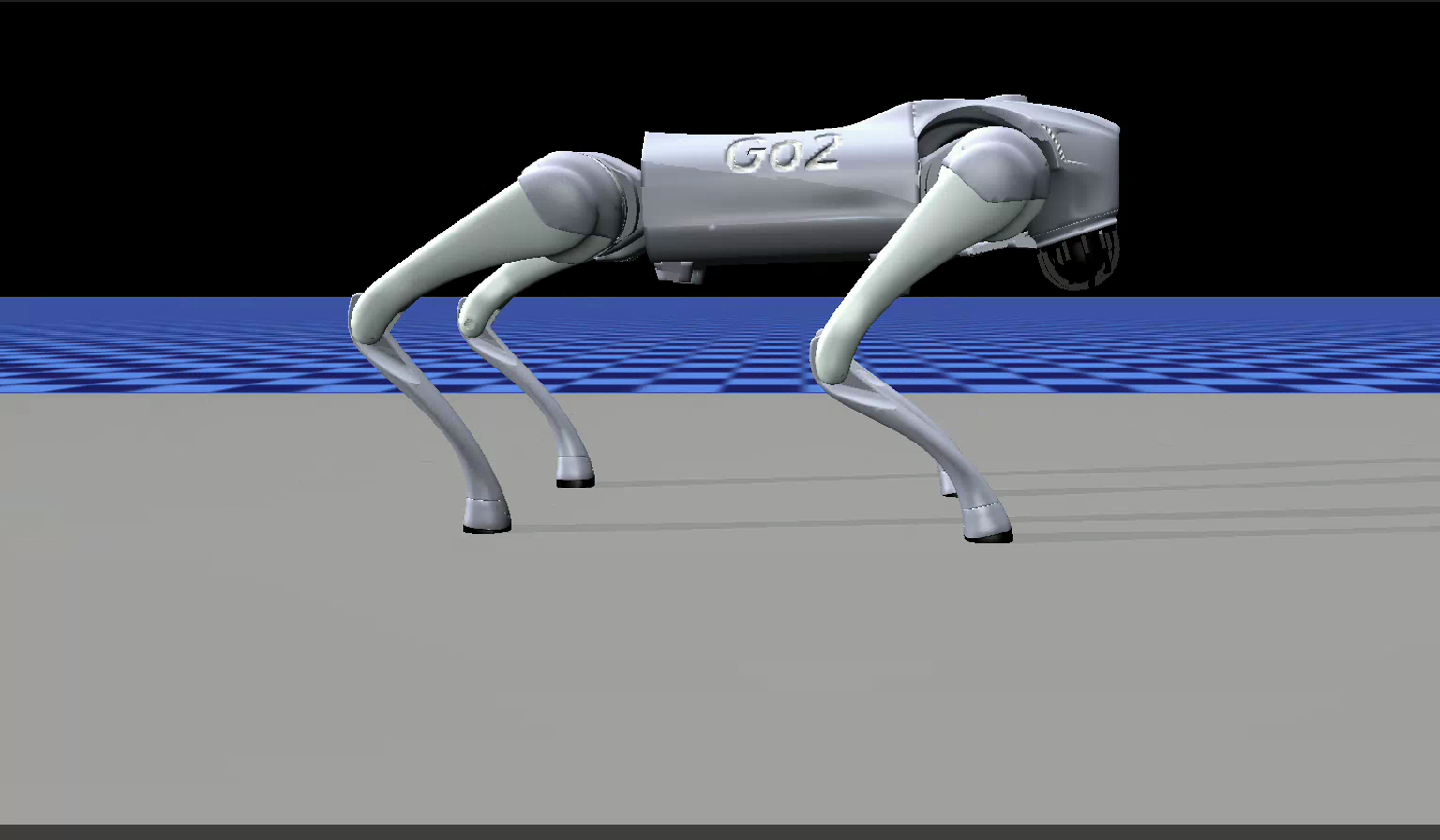}%
        \includegraphics[width = 0.3\textwidth]{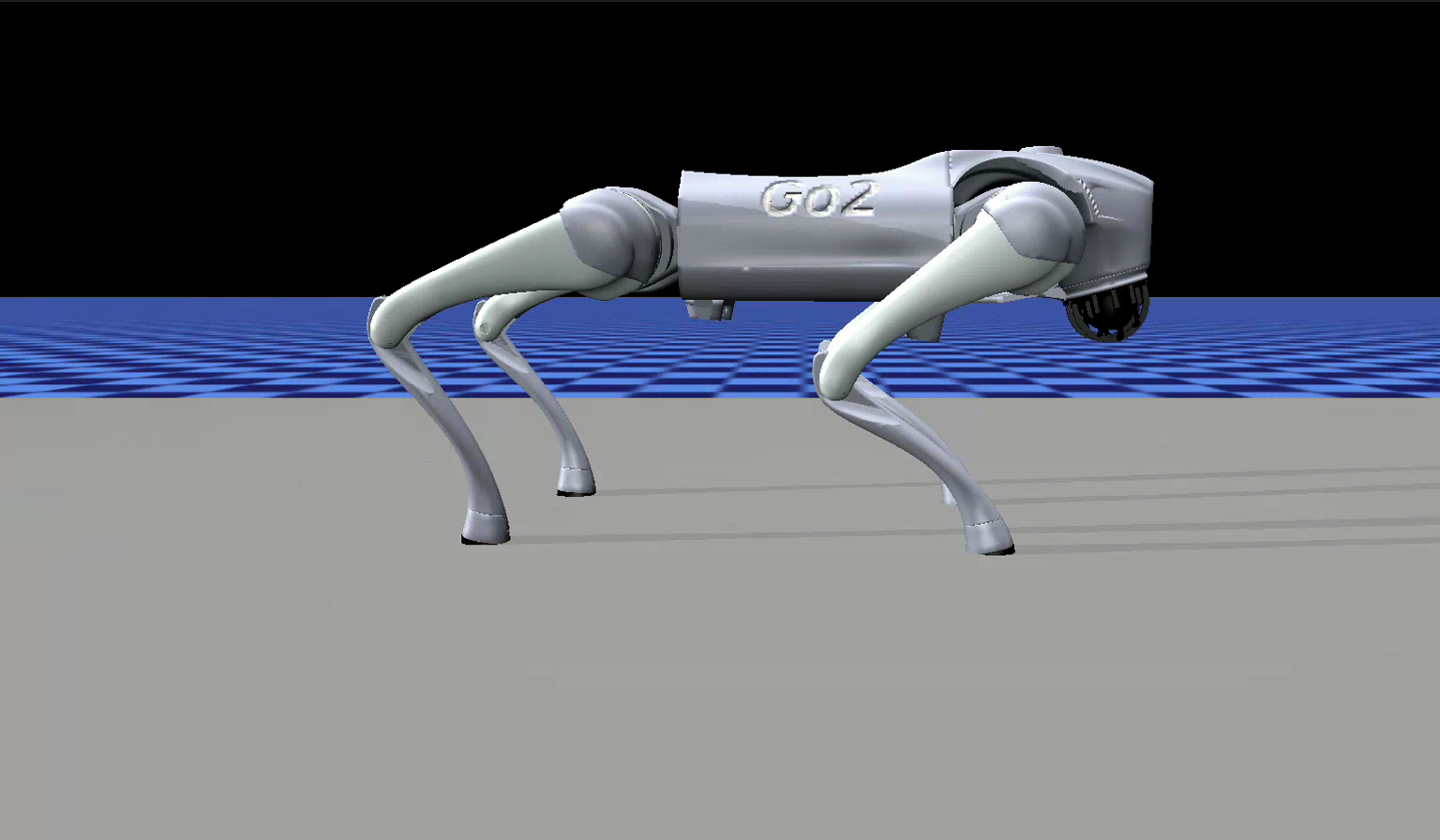}%
        \\
        \includegraphics[width = 0.3\textwidth]{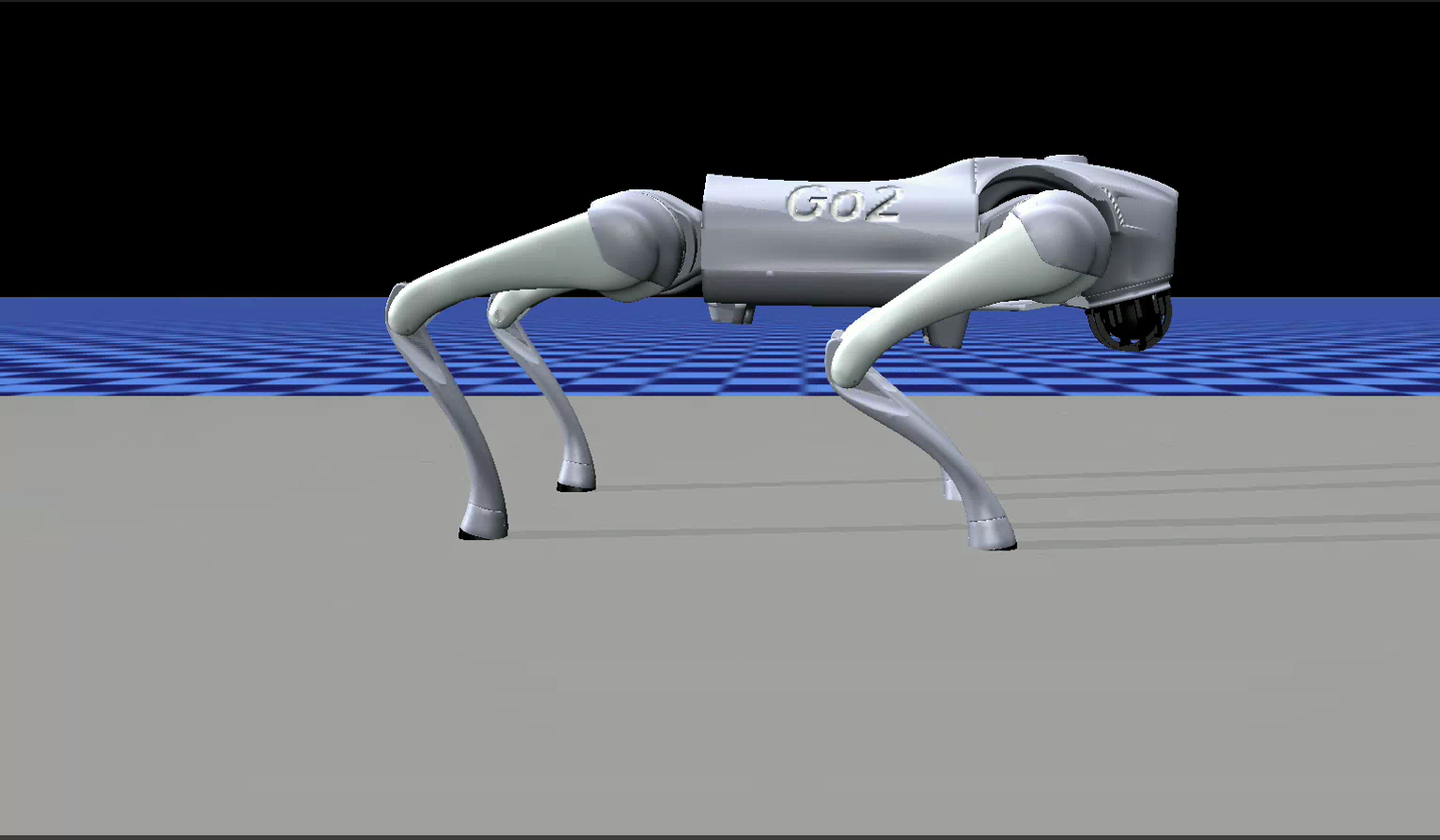}%
        \includegraphics[width = 0.3\textwidth]{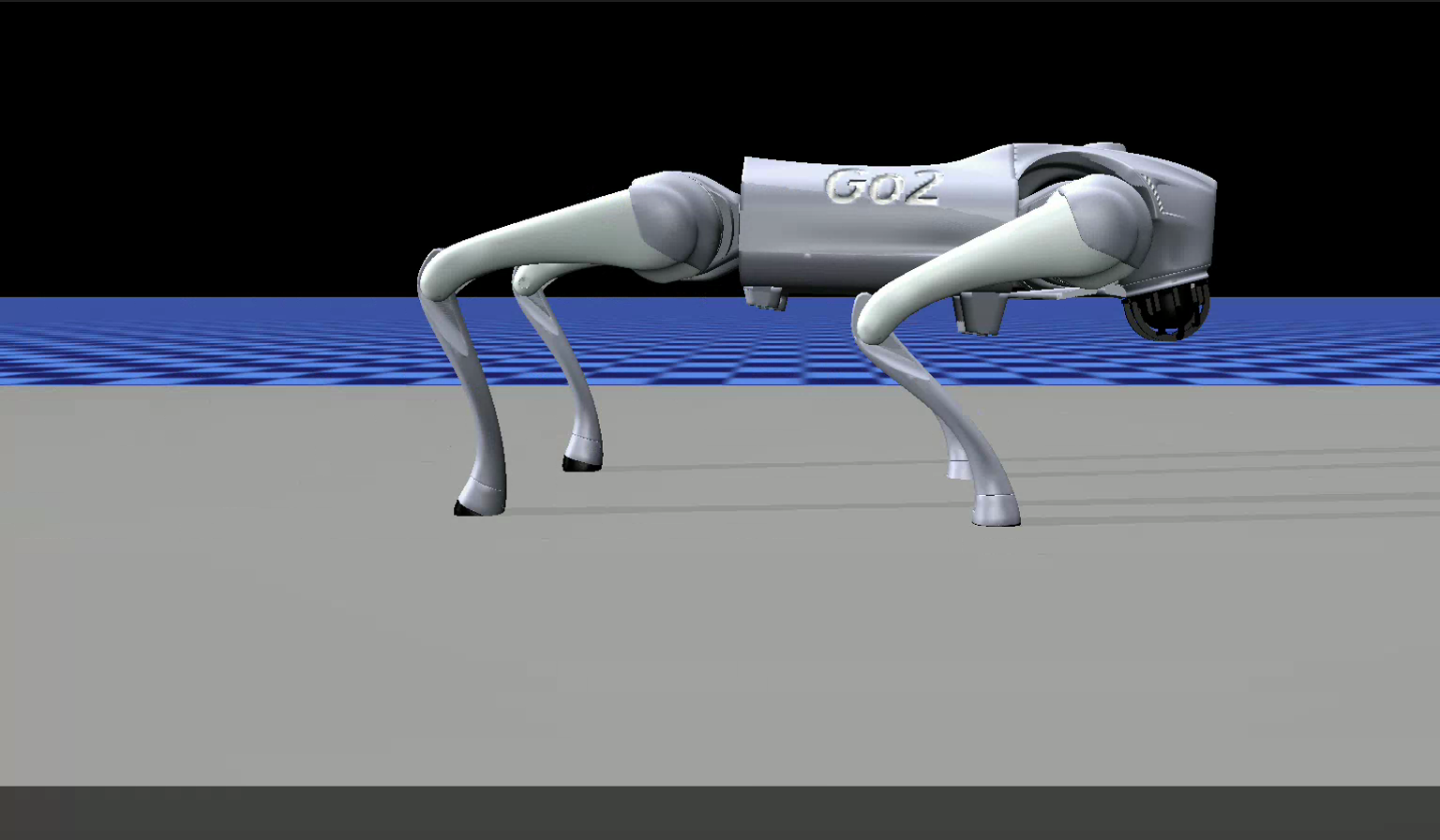}%
        \includegraphics[width = 0.3\textwidth]{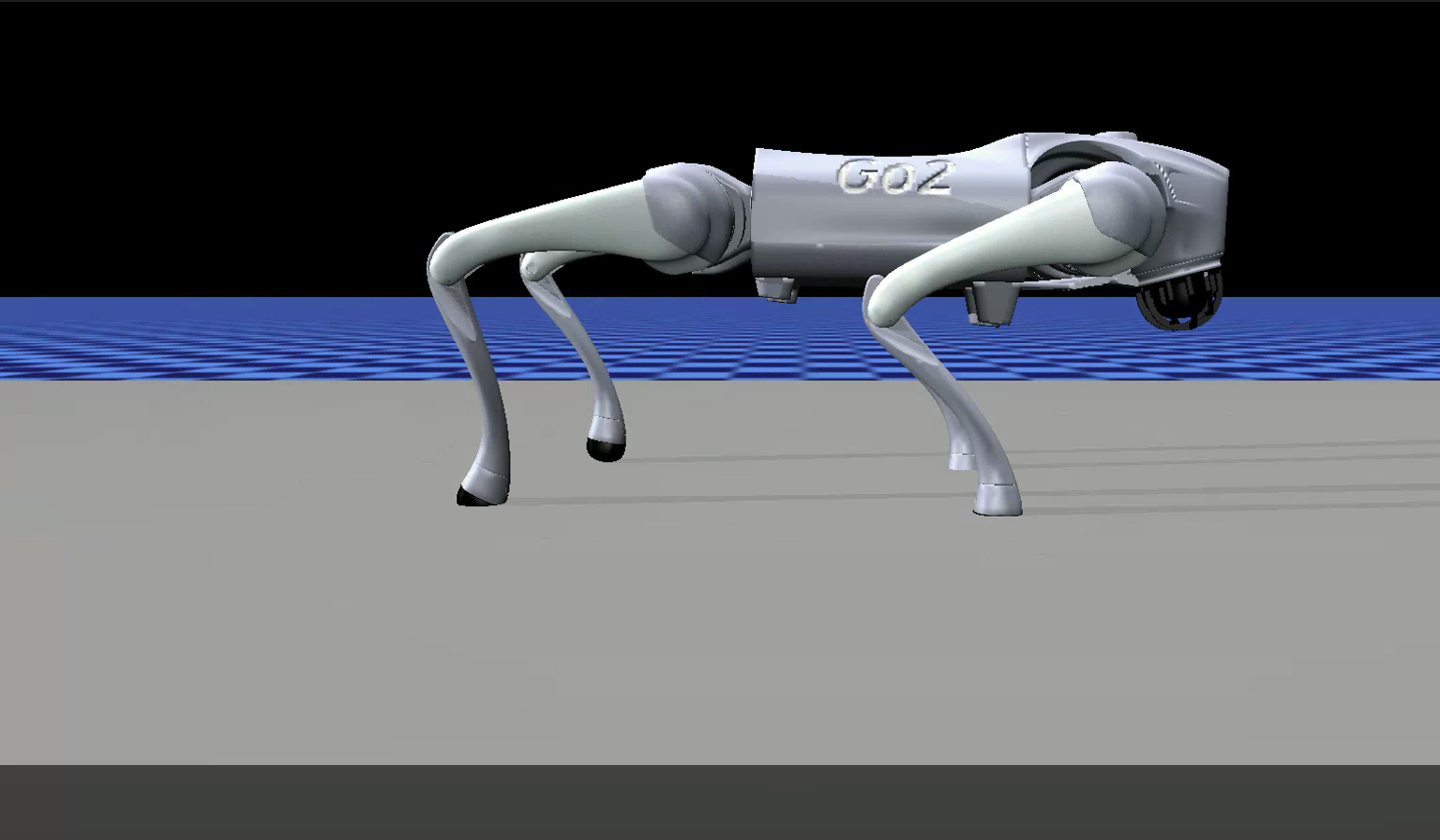}%
    \end{subfigure}
    \hfill
    \begin{subfigure}{0.49\textwidth}
        \centering
        \includegraphics[width = 0.3\textwidth]{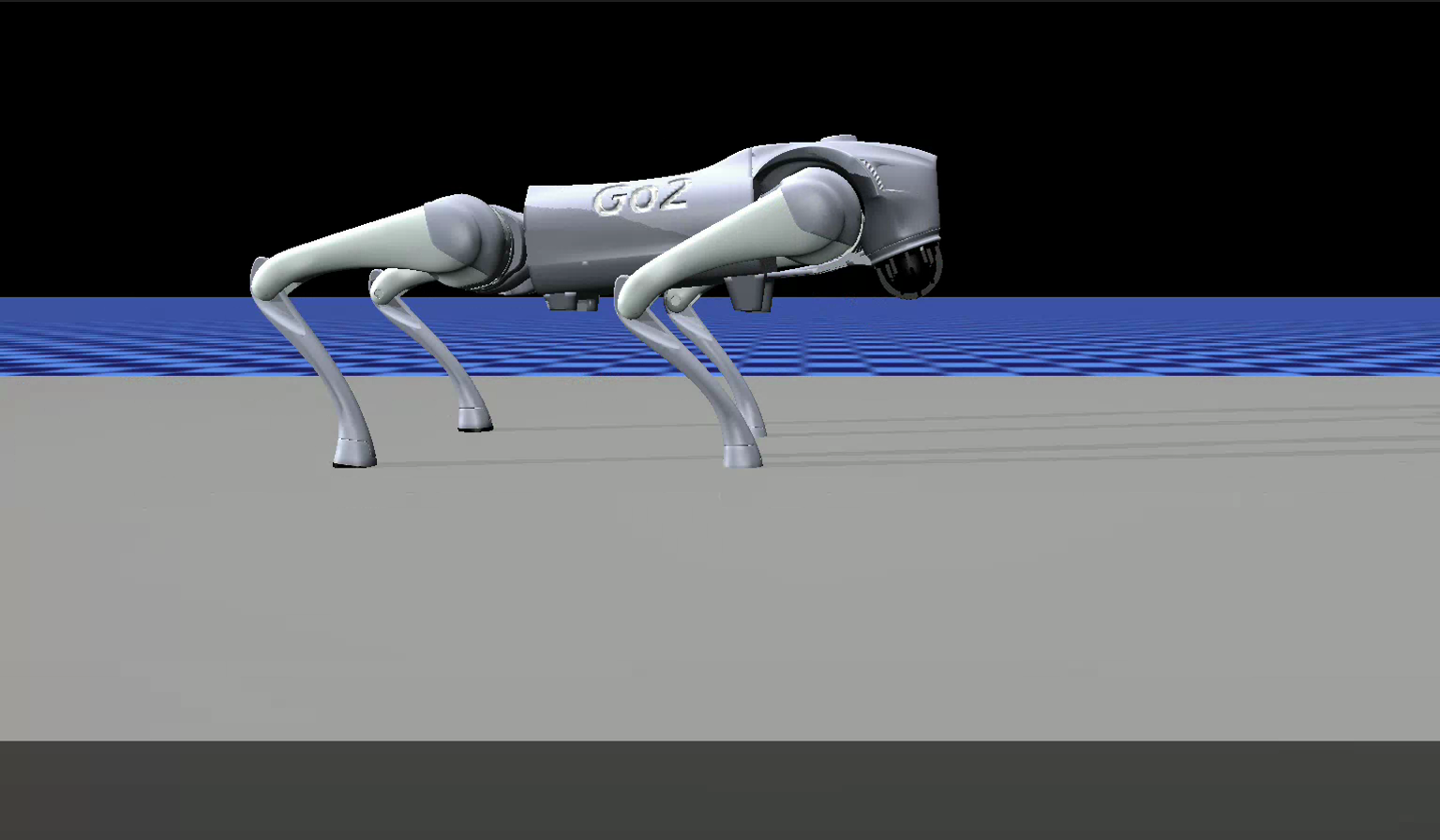}%
        \includegraphics[width = 0.3\textwidth]{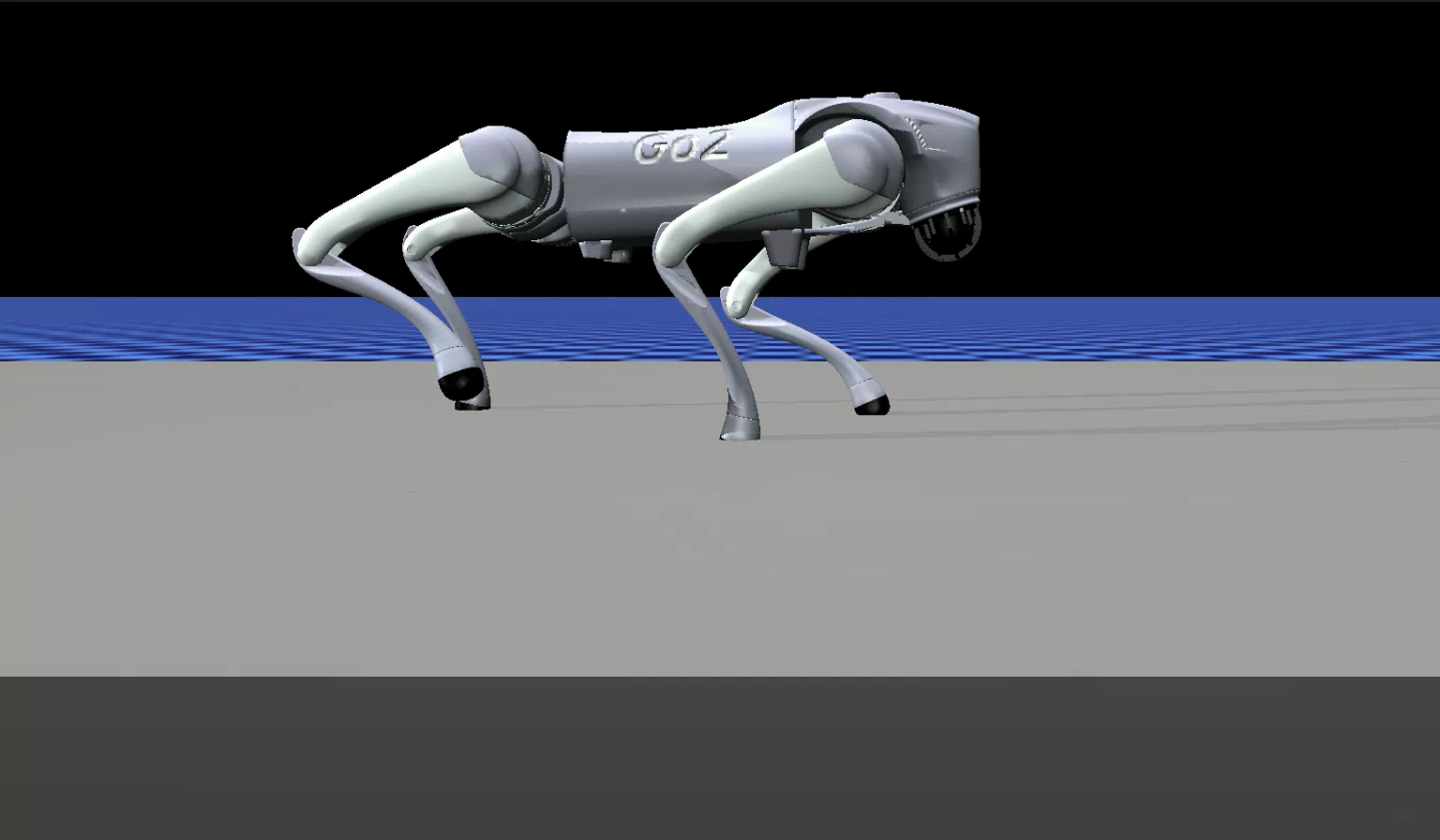}%
        \includegraphics[width = 0.3\textwidth]{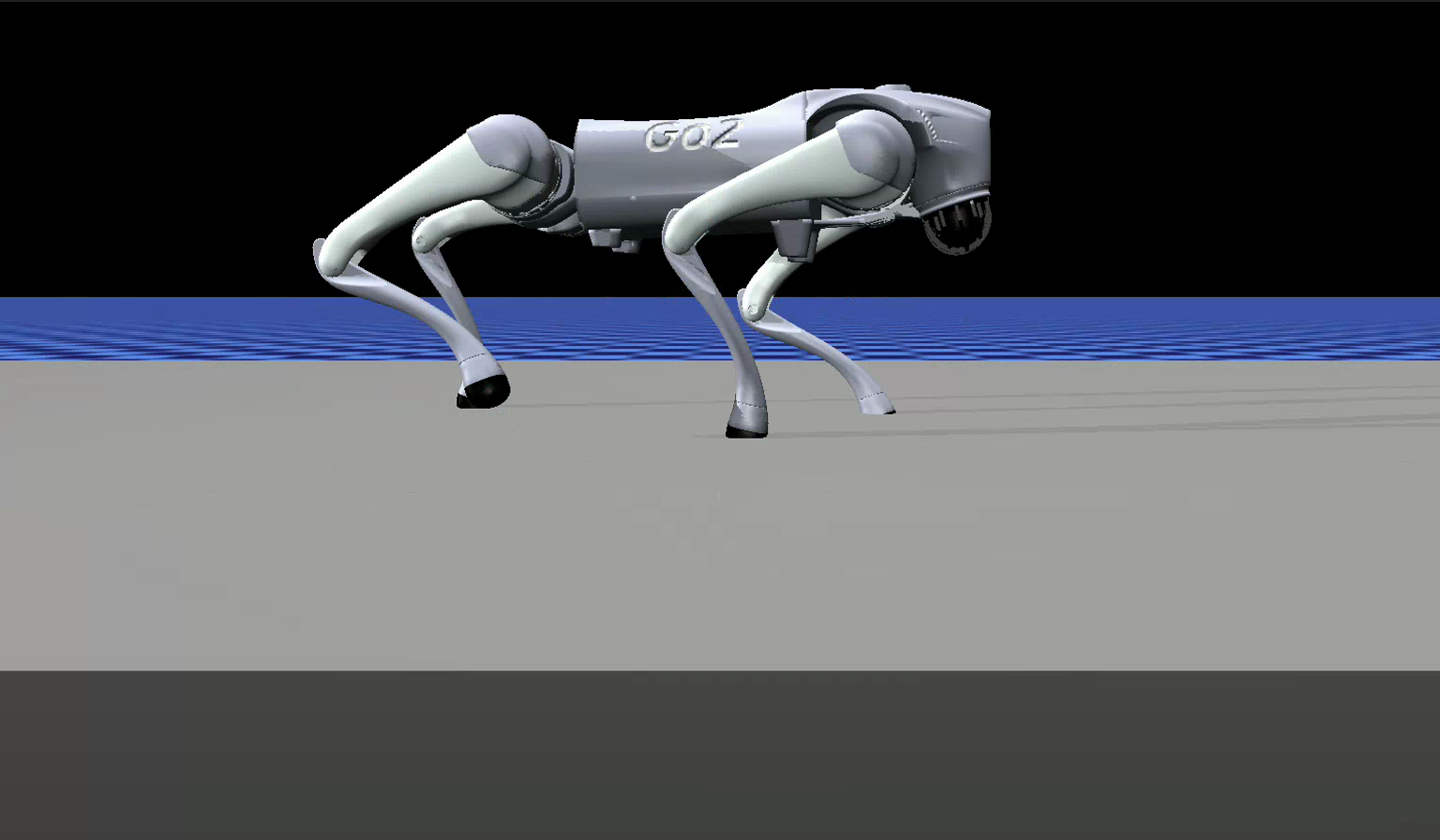}%
        \\
        \includegraphics[width = 0.3\textwidth]{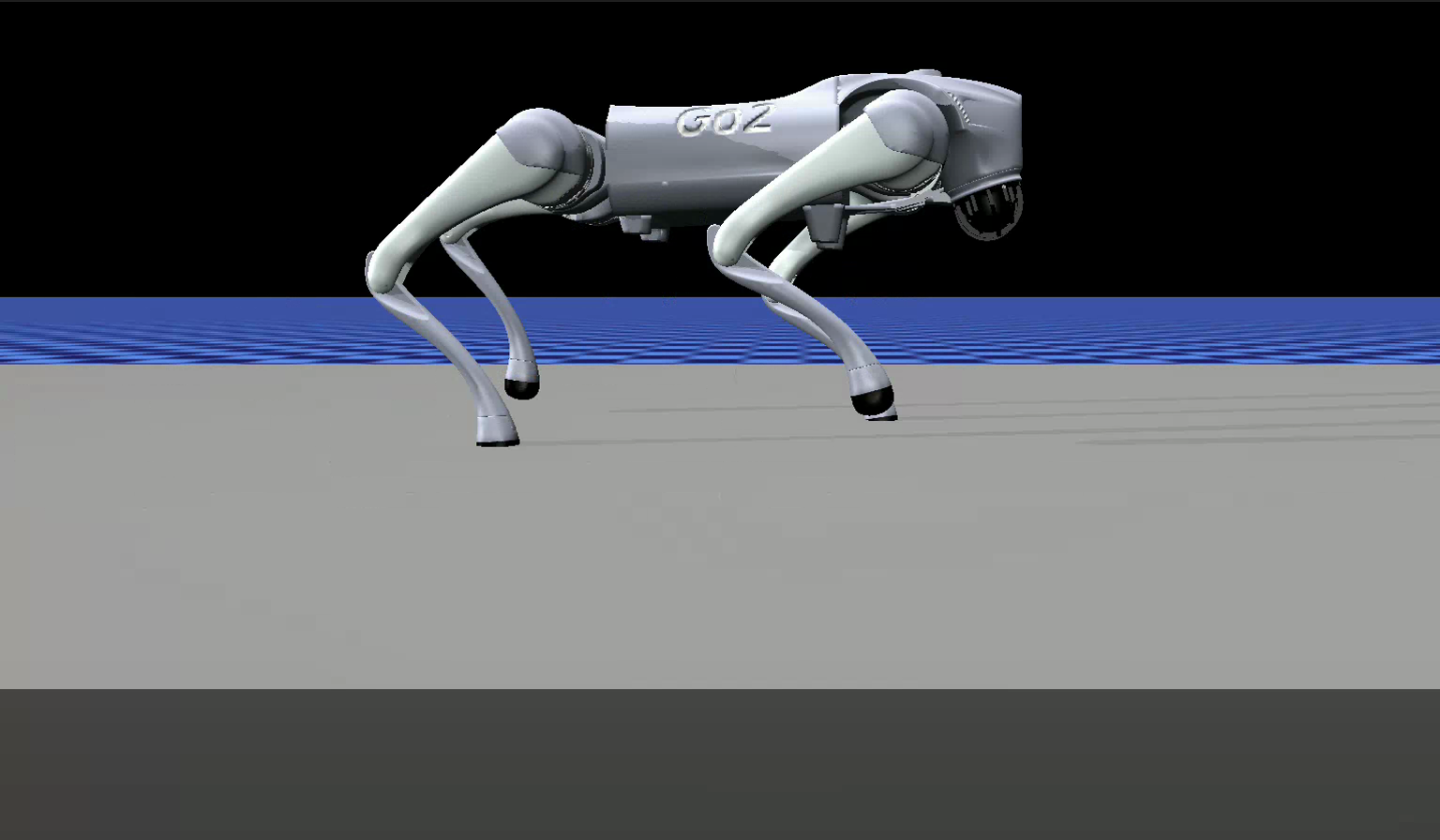}%
        \includegraphics[width = 0.3\textwidth]{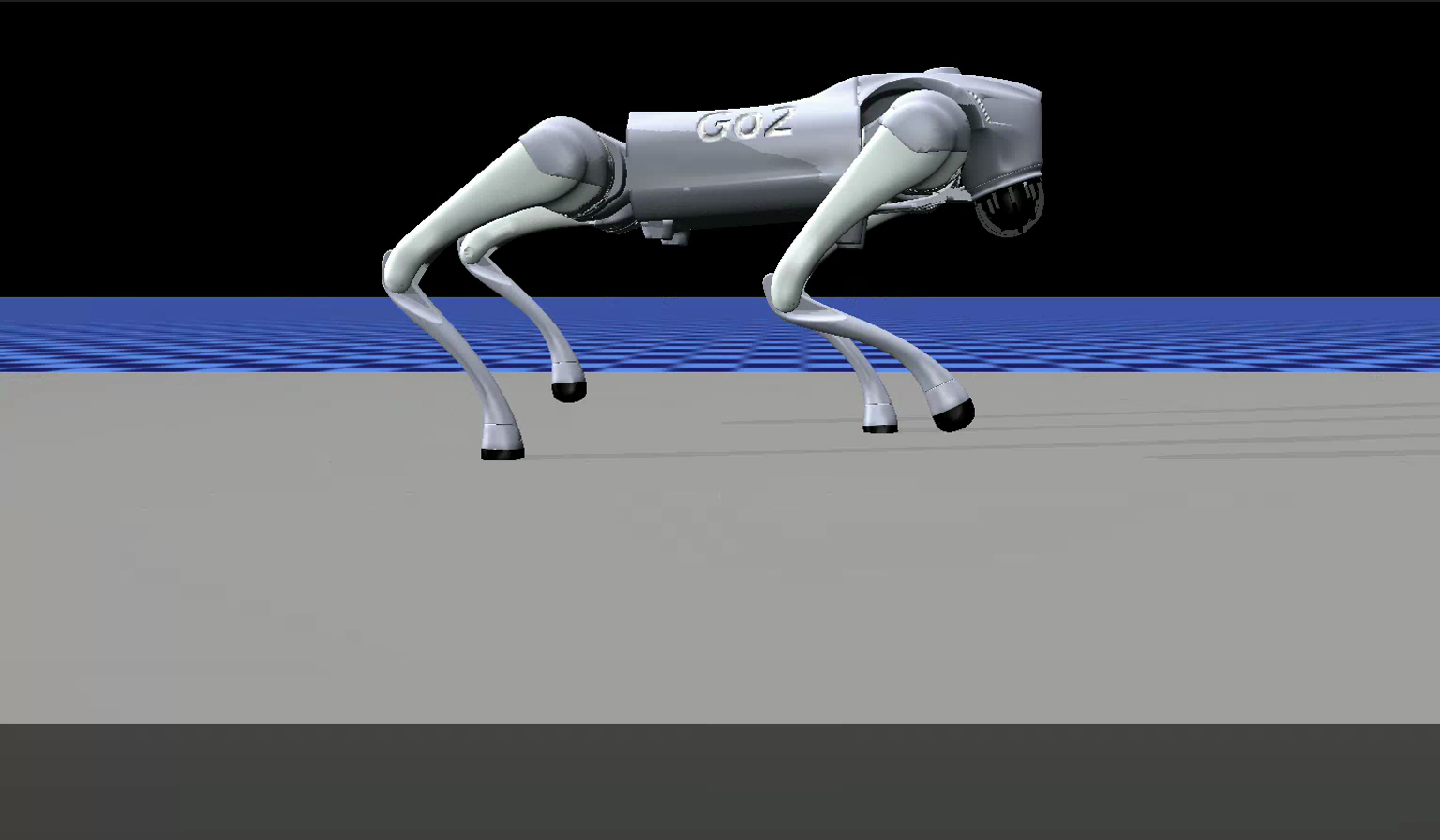}%
        \includegraphics[width = 0.3\textwidth]{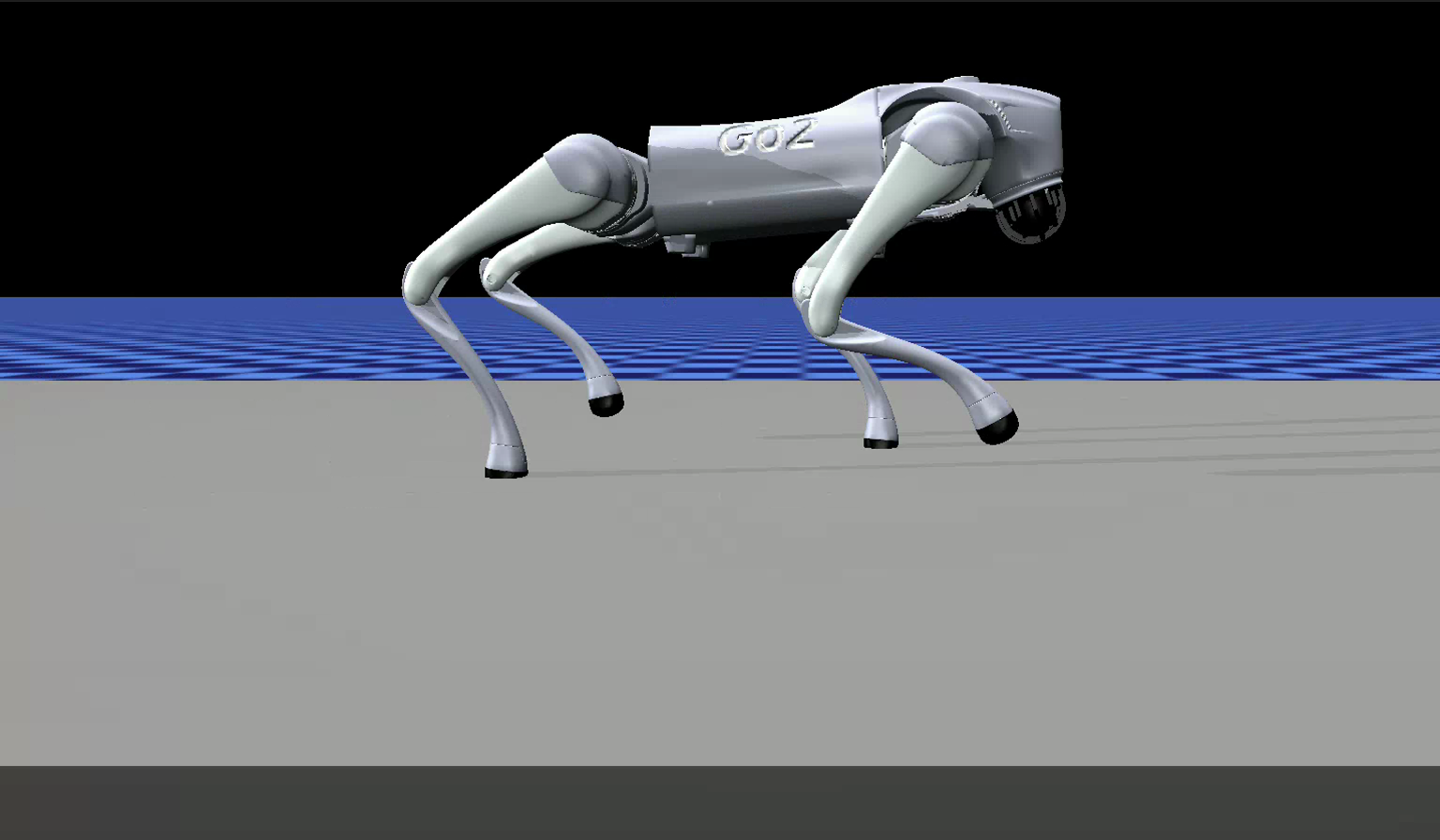}%
        \\
        \includegraphics[width = 0.3\textwidth]{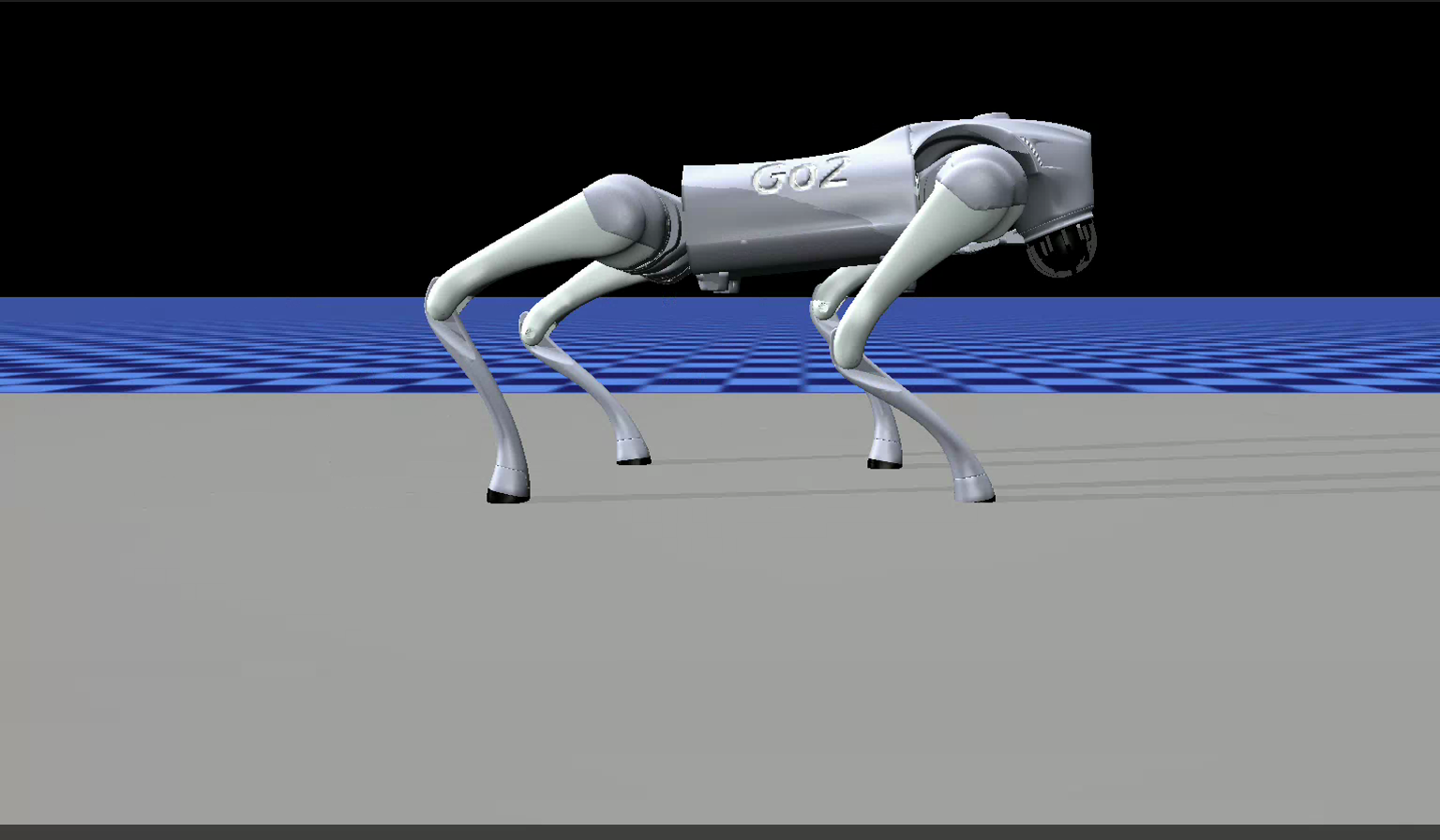}%
        \includegraphics[width = 0.3\textwidth]{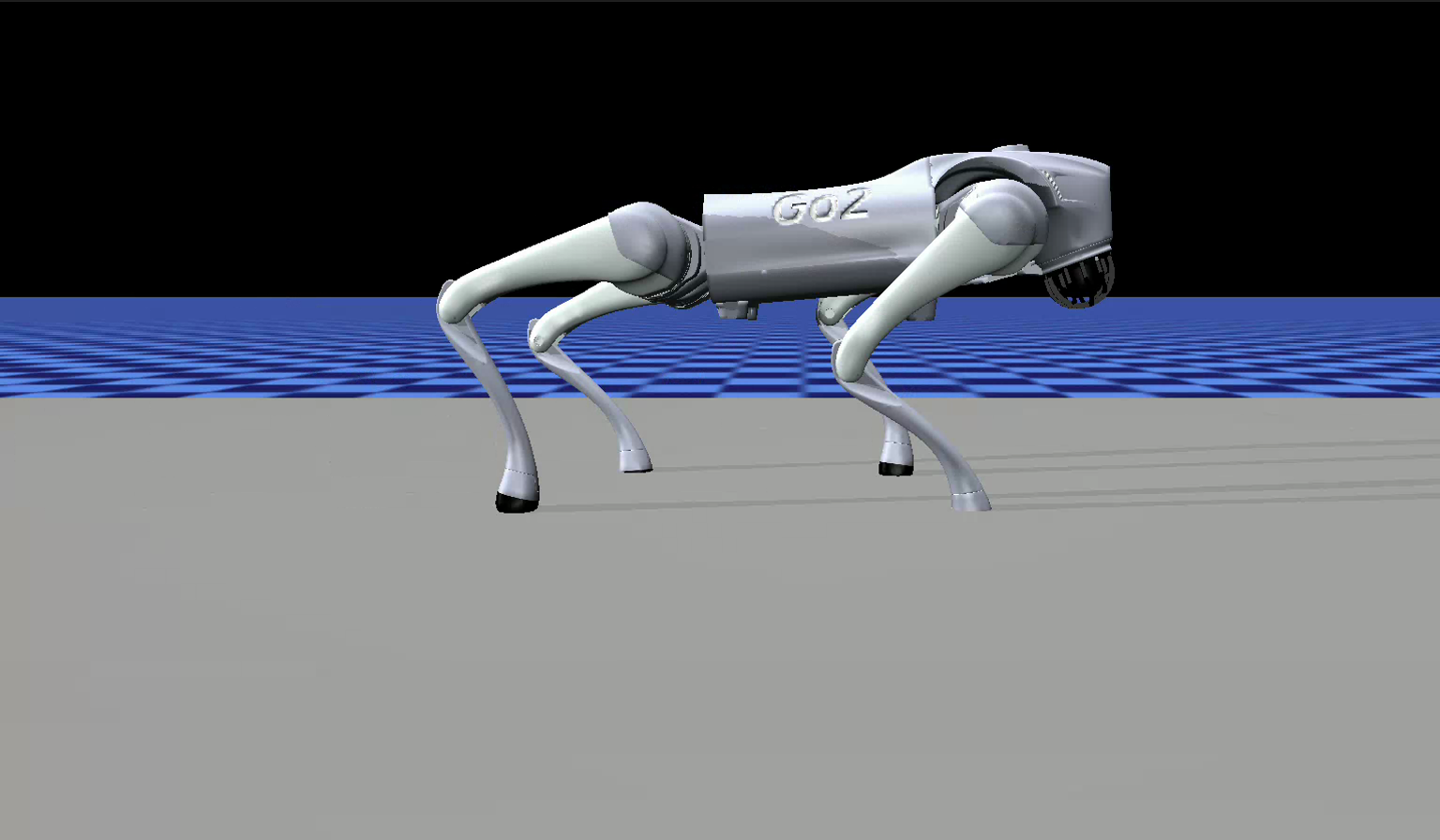}%
        \includegraphics[width = 0.3\textwidth]{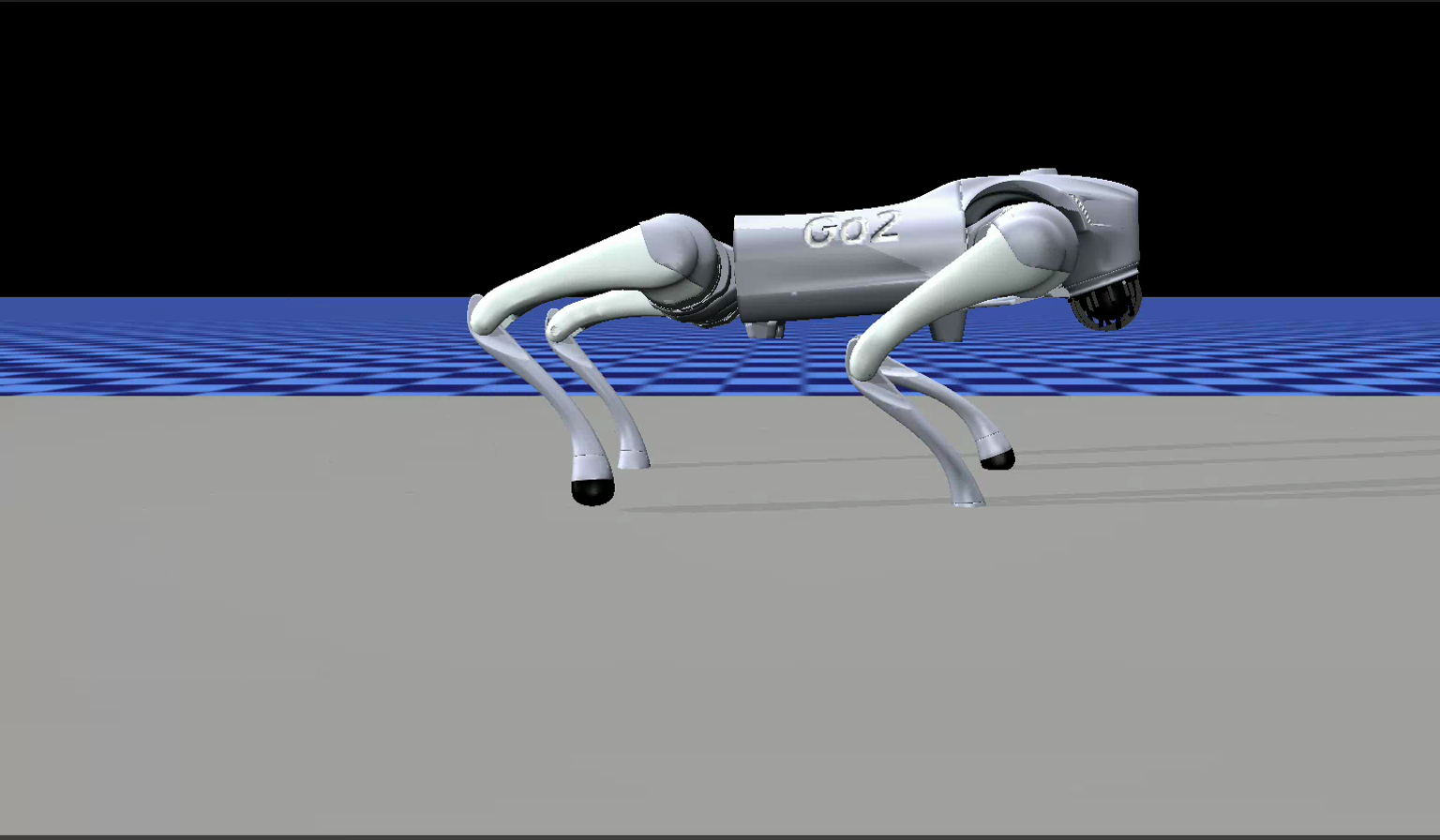}%
    \end{subfigure}
    \caption{
        Snapshots of simulated gait cycles on the oscillating bridge for \textit{bound nos} (left) and \textit{default eb} (right).
        Each set is arranged left to right, row-wise, spanning one full oscillation cycle of the bridge.
    }
    \label{fig.snapshots_combined}
\end{figure*}

To investigate how the different policies cope with the oscillating bridge, we evaluated the movement of the robot's \gls{com} compared to the bridge's surface.
\autoref{fig.CoM_all} shows the $x$, $y$ and $z$ components of the \gls{com} movement of the robot with a fixed velocity command $v_x = 0.5$\,m/s for the \textit{default} gait style under the three training conditions.
When the bridge is oscillating with a frequency of \SI{2.0}{\hertz} and an amplitude $\pm$\SI{0.1}{\meter}, all policies clearly show adaptation in their $z$-movement to the bridge's oscillation with a small phase shift of around $0.3 \pi$, due to the robot's inertia.
The \textit{nos} policy, however, struggles to keep the robot's \gls{com} constant, exhibits more chaotic behavior, and struggles to walk at the commanded velocity, progressing slower over the bridge than the \textit{eb} and \textit{eg} policies. 
Interestingly, the \textit{eb} and \textit{eg} policies show a lateral drift in the $y$-direction, when the bridge is not oscillating, which is not present in the \textit{nos} policy, as it was trained in this scenario.
On the oscillating bridge, the \textit{eb} and \textit{eg} policies are able to move mostly straight ahead, while now the \textit{nos} policy shows a strong drift in the $y$-direction.
This indicates that the \textit{eb} and \textit{eg} policies are more robust to the oscillating bridge but are slightly worse adapted to the non-oscillating surface.
Additionally, we provide snapshots of the simulated gait cycles on the oscillating bridge for two policies, \textit{bound nos} and \textit{default eb}, in \autoref{fig.snapshots_combined}.

In the real-world validation experiments, the different policies were evaluated on the \gls{humvib} structure, with the exception of the \textit{pace} gait.
This exclusion was due to the instability of the \textit{pace} policies, increasing the risk of the robot falling off the structure and potentially damaging itself.
All other policies could be evaluated and showed a decrease in movement speed while passing over the middle section of the structure, where the oscillation were highest.
The \textit{eb} and \textit{eg} policies showed a more stable performance with regard to vertical and lateral movements of the \gls{com}.
Interestingly, although the \textit{bound} gait performed poor in simulation, it was capable of stable locomotion on the oscillating bridge in the real world, hinting at inherent robustness and transferability of the gait to the real system.
While the bridge was pre-oscillated by human operators, the \textit{trot} policies were able to excite the structure, since their step frequency of around \SI{4.0}{\hertz} fell close to the 2nd harmonic of the structure.

\begin{figure}
    \centering
    \includegraphics[width=\linewidth]{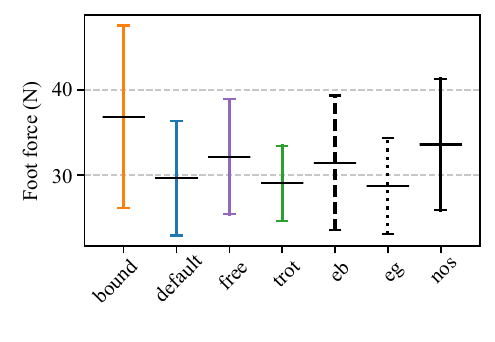}
    \caption{
        Means and standard deviations of the force readouts of the foot sensors of the robot on the oscillating \gls{humvib} bridge in the real world.
    }
    \label{fig.forces}
\end{figure}

To assess the interaction between the robot and the bridge, we measured the mean forces in the feet of the robot.
When oscillating the bridge, the \textit{nos} policies showed the highest forces, while the \textit{eb} policies recorded the lowest ones (\autoref{fig.forces}), indicating the learned adaptation to the oscillations of the real bridge when trained with the simulated model.
Considering the gait styles, \textit{trot} and \textit{default} recorded similar forces, while the \textit{free} policies produced more force, with \textit{bound} topping the scale (\autoref{fig.forces}), due to its inherent leaping motion.
Further analysis of the power usage of the different gaits and styles can be found in appendix \ref{app:power_usage}.

\section{DISCUSSION}
During the learning phase, the \textit{free} gait was the easiest to learn for the \gls{rl} agents.
This is expected, since it imposes no restrictions on the footfall patterns compared to the gait styles found in quadrupeds in nature.
Also, the worse performance of the \textit{bound} and \textit{pace} policies is expected, since they include relatively longer contact times, which were penalized by the learner, and are inherently less stable compared to the \textit{trot} and \textit{default} gaits.
While this study evaluated the performance of 15 different policies in the given scenario, each policy was only trained with one seed, limiting the statistical power.
This could also be part of the reason, why the \textit{pronk} policy reversed to a \textit{trot} behavior during training.
Further optimization of the policies is required to provide a exhaustive comparison of the different gait styles.
Since currently the real-world bridge can only exhibit an eigenfrequency of \SI{2.0}{\hertz} or be rigid, further frequencies could not be tested in our study.
When looking at the different height regulation strategies, the \textit{nos} policies superiority during learning is a product of the simpler training environment, as the robot is not disturbed by oscillations of the bridge.
The performance differences of the policies with respect to the \gls{com} movement in simulation can also be explained by their relative training scenarios.
Since the \textit{eb} and \textit{eg} policies trained on oscillating surfaces, they had a harder time to adapt to the rigid bridge.
However, the oscillation seems to impose so much of a disturbance that the drift of the \textit{nos} policy becomes excessive to a point where it would fall off the bridge after less than \SI{10}{\second}.

While most policies cope well with the setting they used for learning, we can observe that the current black-box modelling approach does not generalize perfectly to situations that have not been learned. A mixed black-box and white-box model approach could be used to overcome this limitation. White-box models are structural models that implement physiological details about the sensorimotor system (e.g., neural circutries,  muscle properties \cite{schumacher2017sensor}, sensor-mechanical couplings \cite{rongala2024import}). This combination of structure models and learning may help to better understand the versatility of humans and animals who are able to traverse over moving grounds benefitting of the underlying neural controller dynamics \cite{van2009robust}.

Future work will extend our research to explore the combination of different ground and surface perturbations with moving obstacles, as well as the integration of high-level planning and visual navigation systems \cite{aditya2025robust} to traverse complex multi-layered terrains with dynamic and adversarial perturbations.

Lastly, while no quadrupedal animals have been tested on the \gls{humvib} bridge yet, future research is planned to understand the biological coping strategies in this scenario, covering the full picture of quadrupedal gaits during vertical ground oscillations \cite{kalveram2009inverse}.

\section{CONCLUSION}
This study demonstrates that locomotion policies trained in simulation on an oscillating surface significantly outperform those developed on rigid terrain when tested on the Unitree Go2 quadruped navigating the \gls{humvib} bridge.
By employing \gls{rl} with the \gls{ppo} algorithm across 15 distinct policies---spanning five gaits and three training conditions---we found that exposure to vertical ground perturbations during training enhances stability and adaptability, a result validated through zero-shot transfer to the physical bridge.
Notably, this work marks the first effort to both simulate and experimentally evaluate a robot's response to such dynamic bridge-induced disturbances, revealing robust gait patterns that withstand real-world instabilities without prior exposure to the testbed.
Our findings underscore the advantage of incorporating dynamic perturbations into simulation environments, advancing the understanding of quadruped locomotion under challenging conditions and paving the way for designing more robust robots capable of operating effectively in dynamic and unpredictable environments.





\appendix
\subsection{Reward function}
\label{app:reward_function}

We split the reward function into three tables: generic terms (\autoref{tab:rew_generic}), base–height definitions (\autoref{tab:rew_height}), and gait‐specific symmetry penalties (\autoref{tab:rew_symmetry}).

\begin{table}[hb!]
  \caption{Generic reward terms.  Here $v_{xyz}$ is the robot’s linear velocity, $\omega_{\mathrm{pitch,roll,yaw}}$ its angular velocity, $\theta_{\mathrm{pitch,roll,yaw}}$ its orientation, $q$ joint positions, $\ddot q$ joint accelerations, $\tau$ joint torques, $a$ actions, $g_f\in\{0,1\}$ foot‐contact flags, $g_f^T$ last foot contact (s), and $n_{\mathrm{collisions}}$ are the number of self-collisions.}
  \label{tab:rew_generic}
  \centering
  \addtolength{\tabcolsep}{-0.4em}
  \begin{tabular}{l l l}
    \toprule
    Term & Coeff. & Equation \\
    \midrule
    XY velocity tracking        & $2.0$  & $\displaystyle \exp\bigl(-\|v_{xy}-\bar v_{xy}\|^2/0.25\bigr)$ \\
    Yaw velocity tracking       & $1.0$  & $\displaystyle \exp\bigl(-|\omega_{\mathrm{yaw}}-\bar v_{\mathrm{yaw}}|^2/0.25\bigr)$ \\
    Z velocity penalty          & $2.0$  & $-\,|v_z|^2$ \\
    Pitch–roll velocity penalty & $0.05$ & $-\,\|\omega_{\mathrm{pitch,roll}}\|^2$ \\
    Pitch–roll position penalty & $0.2$  & $-\,\|\theta_{\mathrm{pitch,roll}}\|^2$ \\
    Joint limits penalty        & $10.0$ & $-\,\bigl(0.9q_{\min}<q<0.9q_{\max}\bigr)$ \\
    Joint accel.\ penalty       & $2.5\times10^{-7}$ & $-\,\|\ddot q\|^2$ \\
    Joint torque penalty        & $2.0\times10^{-4}$ & $-\,\|\tau\|^2$ \\
    Action rate penalty         & $0.01$ & $-\,\|\dot a\|^2$ \\
    Collisions penalty          & $1.0$  & $-\,n_{\mathrm{collisions}}$ \\
    Air‐time penalty            & $0.1$  & $\displaystyle -\sum_f g_f\,(g_f^T-0.5)$ \\
    Height penalty               & $30.0$ & $-\bigl(h(t)-h_{\mathrm{nominal}}\bigr)^2$ \\
    Symmetry penalty             & $0.5$  & see \autoref{tab:rew_symmetry} \\
    \bottomrule
  \end{tabular}
  \vspace{-0.5em}
\end{table}

\begin{table}[hb!]
  \caption{Base–height variable $h(t)$ for the training styles. Here $b_0$ is the nominal (equilibrium) bridge height above ground, 
    $z_b(t)$ instantaneous vertical displacement of the bridge, 
    $g$ gravitational acceleration, 
    $m$ robot’s total mass, 
    $k$ stiffness constant, and 
    $A$ oscillation amplitude of the bridge.}
  \label{tab:rew_height}
  \centering
  \begin{tabular}{l l}
    \toprule
    Style & Definition of $h(t)$ \\
    \midrule
    No oscillation (\textit{nos})    & $h_{\mathrm{nos}}(t)=\mathrm{CoM}_z(t)-b_0$ \\[1ex]
    Equidistant bridge (\textit{eb}) & $h_{\mathrm{eb}}(t)=\mathrm{CoM}_z(t)-b_0 - z_b(t)$ \\[1ex]
    Equidistant ground (\textit{eg}) & $h_{\mathrm{eg}}(t)=\mathrm{CoM}_z(t)-b_0 + \tfrac{g\,m}{k} - \tfrac{A}{2}$ \\
    \bottomrule
  \end{tabular}
\end{table}

\begin{table}[ht!]
  \caption{Gait‐specific symmetry penalties.}
  \label{tab:rew_symmetry}
  \centering
  \begin{tabular}{l l}
    \toprule
    Gait   & Symmetry penalty $r_{\mathrm{sym}}(t)$ \\
    \midrule
    Default
      & $-(\neg g_{\mathrm{fr}}\land\neg g_{\mathrm{fl}})
        +(\neg g_{\mathrm{rr}}\land\neg g_{\mathrm{rl}})$ \\[1ex]

    Trot
      & $-(g_{\mathrm{fr}}\neq g_{\mathrm{rl}})
        +(g_{\mathrm{fl}}\neq g_{\mathrm{rr}})
        +(g_{\mathrm{fr}}=g_{\mathrm{fl}}=g_{\mathrm{rr}}=g_{\mathrm{rl}})$ \\[1ex]

    Pace
      & $-(g_{\mathrm{fr}}\neq g_{\mathrm{rr}})
        +(g_{\mathrm{fl}}\neq g_{\mathrm{rl}})
        +(g_{\mathrm{fr}}=g_{\mathrm{fl}}=g_{\mathrm{rr}}=g_{\mathrm{rl}})$ \\[1ex]

    Bound
      & $-(g_{\mathrm{fr}}\neq g_{\mathrm{fl}})
        +(g_{\mathrm{rr}}\neq g_{\mathrm{rl}})
        +(g_{\mathrm{fr}}=g_{\mathrm{fl}}=g_{\mathrm{rr}}=g_{\mathrm{rl}})$ \\[1ex]

    Pronk
      & $-\bigl(\neg(g_{\mathrm{fr}}=g_{\mathrm{fl}}=g_{\mathrm{rr}}=g_{\mathrm{rl}})\bigr)$ \\[1ex]

    Free
      & No symmetry penalty (disabled) \\
    \bottomrule
  \end{tabular}
\end{table}

\subsection{Gait phase analysis}
\label{app:gait_percentage}

\autoref{tab:gait_phase_percentages} shows how well the learned policies match the contact patterns induced by the reward terms for the different gaits.
Interestingly, \textit{default} and \textit{free} exhibit stance durations close to \textit{trot}, suggesting a \textit{trot}-like pattern arises naturally.
Also the absence of any airborne phase highlights why \textit{pronk} was excluded in further experiments.

\begin{table}[ht]
  \centering
  \caption{Percentage of time each gait policy (\textit{nos} style) exhibited the characteristic stance phases. $\text{Pronk}_g$ and $\text{Pronk}_a$ denote the proportion of time all feet were on the ground and all feet were in the air, respectively.}
  \label{tab:gait_phase_percentages}
  \begin{tabular}{lrrrrrr}
    \toprule
    Gait   & Default & Trot & Pace & Bound & $\text{Pronk}_g$ & $\text{Pronk}_a$ \\
    \midrule
    default & 100.0      & 35.3     & 0.0      & 0.0       & 35.3       & 0.0         \\
    trot    & 100.0      & 74.3     & 0.0      & 0.0       & 8.6        & 0.0         \\
    pace    & 100.0      & 0.0      & 73.3     & 0.0       & 15.2       & 0.0         \\
    bound   & 64.1       & 0.0      & 0.0      & 35.9      & 48.3       & 0.0         \\
    pronk   & 100.0      & 44.7     & 0.0      & 0.0       & 40.1       & 0.0         \\
    free    & 100.0      & 40.1     & 0.0      & 0.0       & 33.7       & 0.0         \\
    \bottomrule
  \end{tabular}
\end{table}

\subsection{Power usage}
\label{app:power_usage}

\autoref{tab:power_used} summarizes the mean estimated power consumption for each gait type, style, and bridge condition. Overall, the \textit{bound} gait demands the most energy, while the \textit{default} gait is the most efficient. Among styles, \textit{eb} uses more power than \textit{eg} and \textit{nos}, and the pre‑oscillated bridge increases consumption slightly over the idle case.

\begin{table}[ht]
    \centering
    \caption{Estimated power averaged over gaits, styles and bridge‐settings.}
    \label{tab:power_used}
    \addtolength{\tabcolsep}{-0.4em}
    \begin{tabular}{lrrrrrrrrr}
        \toprule
        & default & trot   & bound  & free   & eb     & eg     & nos    & idle   & pre    \\
        \midrule
        Watt & 72.11   & 99.64  & 130.65 & 77.64  & 100.21 & 89.89  & 83.57  & 89.98  & 92.84  \\
        \bottomrule
    \end{tabular}
\end{table}


\section*{ACKNOWLEDGMENT}
We sincerely thank the German Research Center for AI (DFKI), Research Department: Systems AI for Robot Learning, for lending us the Unitree Go2 quadruped.


\addtolength{\textheight}{-5.7cm}
\bibliographystyle{IEEEtran}
\bibliography{IEEEabrv,bibliography}



\end{document}